\documentclass[twoside,12pt]{article}

\usepackage[none]{hyphenat}
\usepackage{jmlr2e}
\usepackage{amssymb}
\usepackage{amsmath}
\usepackage{algorithm}
\usepackage{mathtools}
\usepackage{tikz}
\usepackage[noend]{algpseudocode}
\usepackage{caption}
\usepackage{ulem}
\usepackage{glossaries}
\usepackage{glossary-mcols}
\usepackage{xparse}

\makenoidxglossaries
\newglossaryentry{gls_tau} {
  name={\ensuremath{\tau}},
  text={\ensuremath{\tau}},
  description={total number of chronological time points; index for time is $t$}
} 

\newglossaryentry{gls_n} {
  name={\ensuremath{n}},
  text={\ensuremath{n}},
  description={total number of sampled subjects; index for samples is $i$}
}

\newglossaryentry{gls_Yt} {
  name={\ensuremath{Y_i(t)}},
  text={\ensuremath{Y_i(t)}},
  description={time $t$-specific response variable for sample $i$}
}

\newglossaryentry{gls_Wt} {
  name={\ensuremath{W_i(t)}},
  text={\ensuremath{W_i(t)}},
  description={time $t$-specific time-varying covariates for sample $i$}
}

\newglossaryentry{gls_Ot} {
  name={\ensuremath{O_i(t)}},
  text={\ensuremath{O_i(t)}},
  description={time $t$-specific p-dimensional random variable for sample $i$ where $O_i(t) = (W_i(t), Y_i(t))$}
}

\newglossaryentry{gls_X} {
  name={\ensuremath{X_i}},
  text={\ensuremath{X_i}},
  description={vector of baseline covariates for sample $i$}
}

\newglossaryentry{gls_On} {
  name={\ensuremath{O^n}},
  text={\ensuremath{O^n}},
  description={collection of $n$ time-series}
}

\newglossaryentry{gls_P0} {
  name={\ensuremath{P_0^n}},
  text={\ensuremath{P_0^n}},
  description={true probability distribution of $(X^n,O^n)$}
}

\newglossaryentry{gls_Px} {
  name={\ensuremath{P_{0,O_i|X_i}}},
  text={\ensuremath{P_{0,O_i|X_i}}},
  description={true conditional distribution of $O_i$ given $X_i$ (short notation: $P_{0,X_i}$)}
}

\newglossaryentry{gls_M} {
  name={\ensuremath{\mathcal{M}}},
  text={\ensuremath{\mathcal{M}}},
  description={statistical model, such that the true data-generating distribution is an element of it ($P_0^n \in \mathcal{M}$)}
}

\newglossaryentry{gls_Obar} {
  name={\ensuremath{\overline{O}_i(t-1)}},
  text={\ensuremath{\overline{O}_i(t-1)}},
  description={time $t$-specific history of the time-series for sample $i$, where $\overline{O}_i(t-1) = (O_i(1),  \ldots, O_i(t-1))$}
}

\newglossaryentry{gls_Onbar} {
  name={\ensuremath{\overline{O}^n(t-1)}},
  text={\ensuremath{\overline{O}^n(t-1)}},
  description={time $t$-specific history of $n$ time-series, where $\overline{O}^n(t-1) = \{\overline{O}_1(t-1), \ldots,  \overline{O}_n(t-1)\}$}
}

\newglossaryentry{gls_Pcommon} {
  name={\ensuremath{P_{0,O_i}}},
  text={\ensuremath{P_{0,O_i}}},
  description={true common conditional distribution of $O_i$ given $Z_{i}(t-1)$ and $X_i$}
}

\newglossaryentry{gls_Z} {
  name={\ensuremath{Z_i(t-1)}},
  text={\ensuremath{Z_i(t-1)}},
  description={time $t$-specific fixed dimensional summary measure for sample $i$}
}

\newglossaryentry{gls_Psi} {
  name={\ensuremath{\Psi}},
  text={\ensuremath{\Psi}},
  description={parameter mapping, a function from an element of the statistical model $\mathcal{M}$ ($P$) into the parameter space, $\pmb{\Psi}$}
  }
  
\newglossaryentry{gls_psi} {
  name={\ensuremath{\psi}},
  text={\ensuremath{\psi}},
  description={parameter mapping evaluated at a probability distribution in the statistical model, where $\psi \coloneqq \Psi(P)$ for $P \in \mathcal{M}$}
  }
  
\newglossaryentry{gls_estimand} {
  name={\ensuremath{\Psi(P_0^n)(X_i,Z_i(t-1),t)}},
  text={\ensuremath{\Psi(P_0^n)(X_i,Z_i(t-1),t)}},
  description={prediction function evaluated at the truth for the $i^{\text{th}}$ subject and time $t$}
  }  
  
\newglossaryentry{gls_loss} {
  name={\ensuremath{L(\Psi(P))(X_i,Y_i(t),Z_i(t-1))}},
  text={\ensuremath{L(\Psi(P))(X_i,Y_i(t),Z_i(t-1))}},
  description={loss function corresponding to $\Psi(P)$ evaluated for sample $i$ at time $t$; equivalently, we write $L(\psi)(X_i,Y_i(t),Z_i(t-1))$ for $\psi \coloneqq \Psi(P)$}
  }
  
\newglossaryentry{gls_risk} {
  name={\ensuremath{R(P_{0}^n, \psi)}},
  text={\ensuremath{R(P_{0}^n, \psi)}},
  description={true risk of $\psi$, defined as the sum over time and samples of the expected value of the loss with respect to $P_{0}^n$; 
  $R(P_{0}^n, \psi) = \frac{1}{n} \sum_{i=1}^n  \sum_{t=1}^{\tau} E_{P_{0,O_i}} [L(\psi)(X_i,Y_i(t),Z_i(t-1)) | X_i, Z_i(t-1)]$}
  }   
  
\newglossaryentry{gls_psi_0} {
  name={\ensuremath{\psi_{0}}},
  text={\ensuremath{\psi_{0}}},
  description={predictive function corresponding to $\hat{\Psi}(P_{0}^n)$; minimizer over the true risk of all evaluated $\psi$ in the parameter space}
  }  
 
\newglossaryentry{gls_psi_0i} {
  name={\ensuremath{\psi_{0,i}}},
  text={\ensuremath{\psi_{0,i}}},
  description={minimizer over the true risk of all evaluated $\psi$ in the parameter space for sample $i$}
  }   
  
\newglossaryentry{gls_hatPsi} {
  name={\ensuremath{\hat{\Psi}}},
  text={\ensuremath{\hat{\Psi}}},
  description={estimator mapping, a function from the empirical distribution ($P_{n,t}$, at any $t$) to the parameter space, $\pmb{\Psi}$}
  }  

\newglossaryentry{gls_Pnt} {
  name={\ensuremath{P_{n,t}}},
  text={\ensuremath{P_{n,t}}},
  description={empirical distribution of $n$ time series collected until time $t$}
} 

\newglossaryentry{gls_psi_nt} {
  name={\ensuremath{\psi_{n,t}}},
  text={\ensuremath{\psi_{n,t}}},
  description={predictive function corresponding to $\hat{\Psi}(P_{n,t})$; $\hat{\Psi}(P_{n,t})$ maps $(X_i,Z_i(t-1),t)$ into a time- and subject-specific outcome, $Y_{i}(t)$}
  }
  
\newglossaryentry{gls_v} {
  name={\ensuremath{v}},
  text={\ensuremath{v}},
  description={fold number corresponding to the used cross-validation scheme}
  } 
  
\newglossaryentry{gls_C} {
  name={\ensuremath{C(i,s,\cdot)}},
  text={\ensuremath{C(i,s,\cdot)}},
  description={at minimum, time $s$- and unit $i$-specific record where $C(i,s,\cdot) = (X_i, Z_i(s-1), Y_i(s),\cdot)$}
} 
  
\newglossaryentry{gls_Bn} {
  name={\ensuremath{B_t^v(i,s,\cdot)}},
  text={\ensuremath{B_t^v(i,s,\cdot)}},
  description={$v$-fold assignment of unit $i$ at time point $s$ for split $B_t^v$ trained on data until time $t$. If dynamic enrollment setting, $B_t^v$ also includes starting time $E_i$}
} 

\newglossaryentry{gls_Pnt_0} {
  name={\ensuremath{P_{n,t}^0}},
  text={\ensuremath{P_{n,t}^0}},
  description={empirical distribution of the training sample until time $t$}
} 

\newglossaryentry{gls_Pnt_1} {
  name={\ensuremath{P_{n,t}^1}},
  text={\ensuremath{P_{n,t}^1}},
  description={empirical distribution of the validation sample until time $t$}
} 

\newglossaryentry{gls_nt_0} {
  name={\ensuremath{n_{t}^0}},
  text={\ensuremath{n_{t}^0}},
  description={number of observations in the training set for split $B_t$, over all folds $v$ until time $t$}
} 

\newglossaryentry{gls_nt_1} {
  name={\ensuremath{n_{t}^1}},
  text={\ensuremath{n_{t}^1}},
  description={number of observations in the validation set for split $B_t$, over all folds $v$ until time $t$}
}

\newglossaryentry{gls_Bt_0} {
  name={\ensuremath{\mathcal{B}_{t,v}^0}},
  text={\ensuremath{\mathcal{B}_{t,v}^0}},
  description={set of all $(i,s)$ indexes in the training set until time $t$ for fold $v$}
} 

\newglossaryentry{gls_Bt_1} {
  name={\ensuremath{\mathcal{B}_{t,v}^1}},
  text={\ensuremath{\mathcal{B}_{t,v}^1}},
  description={set of all $(i,s)$ indexes in the validation set until time $t$ for fold $v$}
} 

\newglossaryentry{gls_hatPsi_k} {
  name={\ensuremath{\hat{\Psi}_k}},
  text={\ensuremath{\hat{\Psi}_k}},
  description={estimator mapping for candidate algorithm $k$, a function from the empirical distribution ($P_{n,t}$, at any $t$) to the parameter space, $\pmb{\Psi}$}
  } 

\newglossaryentry{gls_psi_ntk_0} {
  name={\ensuremath{\psi_{n,t,k}^0}},
  text={\ensuremath{\psi_{n,t,k}^0}},
  description={predictive function corresponding to $\hat{\Psi}_k(P_{n,t}^0)$ based on the candidate algorithm $k$ and trained on the training set $P_{n,t}^0$ until time $t$}
  } 

\newglossaryentry{gls_risk_cv} {
  name={\ensuremath{R_{CV}(P_{n,t}^1, \hat{\Psi}_k(\cdot))}},
  text={\ensuremath{R_{CV}(P_{n,t}^1, \hat{\Psi}_k(\cdot))}},
  description={online cross-validated risk for candidate estimator $k$ defined as an empirical average of the loss with respect to $P_{n,t}^1$; in particular, we write $R_{CV}(P_{n,t}^1, \hat{\Psi}_k(\cdot)) = \sum_{j=1}^{t} \sum_{v=1}^V \sum_{(i,s) \in \mathcal{B}_{j,v}^1} L(\psi_{n,j,k}^0)(C(i,s))$}
  } 
  
\newglossaryentry{gls_risk_cv_true} {
  name={\ensuremath{R_{CV}(P_{0}^n, \hat{\Psi}_k(\cdot))}},
  text={\ensuremath{R_{CV}(P_{0}^n, \hat{\Psi}_k(\cdot))}},
  description={the true online cross-validated risk for candidate estimator $k$ defined as the expected value of the loss with respect to $P_{0}^n$; in particular, we write $R_{CV}(P_{0}^n, \hat{\Psi}_k(\cdot)) = \sum_{j=1}^{t} \sum_{v=1}^V \sum_{(i,s) \in \mathcal{B}_{j,v}^1} E_{P_{0,O_i}}[L(\psi_{n,j,k}^0)(C(i,s)) | X_i, Z_i(s-1)]$}
  }   
 
\newglossaryentry{gls_knt} {
  name={\ensuremath{k_{n,t}}},
  text={\ensuremath{k_{n,t}}},
  description={learner that minimizes the online cross-validated risk, the discrete online cross-validation selector}
}   
  
\newglossaryentry{gls_psi_discrete_0} {
  name={\ensuremath{\psi_{n, t, k_{n,t}}^0}},
  text={\ensuremath{\psi_{n, t, k_{n,t}}^0}},
  description={predictive function corresponding to $\hat{\Psi}_{k_{n,t}}(P_{n,t}^0)$, based on the discrete online cross-validation selector $k_{n,t}$ and trained on the training set $P_{n,t}^0$ until time $t$}
  } 

\newglossaryentry{gls_d0_i} {
  name={\ensuremath{d_{0,t}(\psi_{n,t,k,i}, \psi_{0,i})}},
  text={\ensuremath{d_{0,t}(\psi_{n,t,k,i}, \psi_{0,i})}},
  description={loss-based dissimilarity for sample $i$ between $k$ estimator $\hat{\Psi}_k$ evaluated until time $t$ and $\psi_{0,i}$ with respect to $P_{0,O_i}$}
} 

\newglossaryentry{gls_d0} {
  name={\ensuremath{d_{0,t}(\psi_{n,t,k}, \psi_0)}},
  text={\ensuremath{d_{0,t}(\psi_{n,t,k}, \psi_0)}},
  description={average of $i$-specific loss-based dissimilarities over all samples evaluated until time $t$}
}

\newglossaryentry{gls_knt_oracle} {
  name={\ensuremath{\overline{k}_{n,t}}},
  text={\ensuremath{\overline{k}_{n,t}}},
  description={oracle selector among discrete set of candidate estimators $\hat{\Psi}_k$ based on $P_{n,t}$.}
} 

\newglossaryentry{gls_hatPsi_alpha} {
  name={\ensuremath{\hat{\Psi}_{\alpha}}},
  text={\ensuremath{\hat{\Psi}_{\alpha}}},
  description={estimator mapping, a function from the empirical distribution ($P_{n,t}$, at any $t$) generating an ensemble of $K$ estimators $(\hat{\Psi}_1, \ldots, \hat{\Psi}_K)$ indexed by a vector of coefficients $\alpha$ $(\alpha_1, \ldots, \alpha_K)$}
  } 

\newglossaryentry{gls_psi_nta_0} {
  name={\ensuremath{\psi_{n,t,\alpha}^0}},
  text={\ensuremath{\psi_{n,t,\alpha}^0}},
  description={predictive function corresponding to $\hat{\Psi}_{\alpha}(P_{n,t}^0)$ based on the ensemble of $K$ candidate estimators with weights $\alpha$, trained on the training set $P_{n,t}^0$ until time $t$}
  } 
  
\newglossaryentry{gls_alpha} {
  name={\ensuremath{\alpha_{n,t}}},
  text={\ensuremath{\alpha_{n,t}}},
  description={weight for the ensemble of candidate estimators $\hat{\Psi}_k$ based on $P_{n,t}$ that minimizes the online cross-validated risk} 
} 

\newglossaryentry{gls_alpha_oracle} {
  name={\ensuremath{\overline{\alpha}_{n,t}}},
  text={\ensuremath{\overline{\alpha}_{n,t}}},
  description={oracle selector among weights of the ensemble of candidate estimators $\hat{\Psi}_k$ based on $P_{n,t}$}
} 

\newglossaryentry{gls_psi_ntk} {
  name={\ensuremath{\psi_{n,t,k}}},
  text={\ensuremath{\psi_{n,t,k}}},
  description={predictive function corresponding to $\hat{\Psi}_k(P_{n,t})$ based on the candidate algorithm $k$}
  } 
 
\newglossaryentry{gls_E} {
  name={\ensuremath{E_i}},
  text={\ensuremath{E_i}},
  description={chronological entry time for time-series $i$. We assume a natural ordering for all $E_i$, with $0 \leq E_1 \leq \ldots \leq E_n$ even if multiple samples enroll around the same time}
}

\newglossaryentry{gls_Mi} {
  name={\ensuremath{M_i}},
  text={\ensuremath{M_i}},
  description={length of the time-series $i$ from the start of follow-up to the last collected time point}
}

\newglossaryentry{gls_T} {
  name={\ensuremath{T_i}},
  text={\ensuremath{T_i}},
  description={chronological exit time for time-series $i$}
}

\newglossaryentry{gls_hi} {
  name={\ensuremath{h_i(t)}},
  text={\ensuremath{h_i(t)}},
  description={one-to-one mapping from $t$-chronological time to $m$-individual time; note that $h_i(E_i) = 0$ and $h_i(T_i) = M_i$; writing just $h(t)$ denotes a general function which can take a vector of start times}
}

\newglossaryentry{gls_Oi_ds} {
  name={\ensuremath{O^{'}_i(m)}},
  text={\ensuremath{O^{'}_i(m)}},
  description={time $m$-specific p-dimensional variable, where $O^{'}_i(m) = (W^{'}_i(m), Y^{'}_i(m))$ and $O^{'}_i(m) = O_i(h_i(t)) = (W_i(h_i(t)), Y_i(h_i(t)))$; in order to relate $O^{'}_i(m)$ (time-series specific time) to $O_i(h_i(t))$ (chronological time), we need the current chronological time $t$ and the subject's entry time, $E_i$}
}

\newglossaryentry{gls_Oi_h} {
  name={\ensuremath{O_i(h_i(t))}},
  text={\ensuremath{O_i(h_i(t))}},
  description={chronological time $t$-specific p-dimensional variable, where $O_i(h_i(t)) = O^{'}_i(m) = (W_i(h_i(t)), Y_i(h_i(t)))$ and $h_i(t) = t - E_i$; in order to relate $O_i(h_i(t))$ (chronological time) to $O^{'}_i(m)$ (time-series specific time), we need the current chronological time $t$ and the subject's entry time, $E_i$}
}

\newglossaryentry{gls_nt} {
  name={\ensuremath{n(t)}},
  text={\ensuremath{n(t)}},
  description={total number of subjects in the study at chronological time $t$}
}

\newglossaryentry{gls_nmt} {
  name={\ensuremath{n_m(t)}},
  text={\ensuremath{n_m(t)}},
  description={total number of subjects in the study with $m$ points by chronological time $t$}
}

\newglossaryentry{gls_Fmt} {
  name={\ensuremath{F^{n(t)}(t)}},
  text={\ensuremath{F^{n(t)}(t)}},
  description={collection of $n(t)$ observed time-series enrolled at or before time point $t$; we write $F^{n(t)}(t) \equiv \{O_i(h_i(t)): \forall i \text{ where }E_i \leq t\}$}
}

\newglossaryentry{gls_Fbar} {
  name={\ensuremath{\overline{F}(t-1)}},
  text={\ensuremath{\overline{F}(t-1)}},
  description={time $t$-specific history of the available series with start time before chronological time $t$, where $\overline{F}(t-1) = \overline{O}(h(t-1)) = (F(0), \ldots, F(t-1))$; $\overline{F}_i(t-1)$ denotes the sample $i$ time $t$-specific history with start time before chronological time $t$}
}

\newglossaryentry{gls_Ch} {
  name={\ensuremath{C_h(i,s,E_i,T_i)}},
  text={\ensuremath{C_h(i,s,E_i,T_i)}},
  description={time $h_i(s)$- and unit $i$-specific record where $C_h(i,s,E_i,T_i) = (X_i, Z_i(h_i(s-1)), Y_i(h(s)),E_i,T_i)$}
} 

\newglossaryentry{gls_I} {
  name={\ensuremath{\mathcal{I}}},
  text={\ensuremath{\mathcal{I}}},
  description={set of all samples $i$ with start date before current chronological time $t$ $(E_i \leq t)$ with data still being collected $(t \leq T_i)$}
} 

\newglossaryentry{gls_risk_m} {
  name={\ensuremath{R_m(P_{0}, \psi)}},
  text={\ensuremath{R_m(P_{0}, \psi)}},
  description={true risk of $\psi$ for a $m$-time-point-specific prediction, defined as the sum over time and available samples of the expected value of the loss with respect to $P_{0}^n$; 
  $R_m(P_{0}, \psi) = \sum_{t=1}^{\tau} \sum_{i \in \mathcal{I}}  \mathbb{I}(h_i(t) = m) E_{P_{0,O_i}} [L(\psi)(C_h(i,t,E_i,T_i)) |X_i, Z_i(h_i(t-1))] $
  }
  }

\newglossaryentry{gls_risk_cv_m} {
  name={\ensuremath{R_{CV,m}(P_{n,t}^1, \hat{\Psi}_k(\cdot))}},
  text={\ensuremath{R_{CV,m}(P_{n,t}^1, \hat{\Psi}_k(\cdot))}},
  description={online cross-validated risk for candidate estimator $k$ defined as an empirical average of the loss with respect to $P_{n,t}^1$ at individual time $m$}
  } 
  
\newglossaryentry{gls_kntm} {
  name={\ensuremath{k_{n,t,m}}},
  text={\ensuremath{k_{n,t,m}}},
  description={learner that minimizes the online cross-validated risk for individual time $m$, the $m$-discrete online cross-validation selector
  }
}  

\newglossaryentry{gls_alpha_ntm} {
  name={\ensuremath{\alpha_{n,t,m}}},
  text={\ensuremath{\alpha_{n,t,m}}},
  description={weight for the ensemble of candidate estimators $\hat{\Psi}_k$ based on $P_{n,t}$ for individual time $m$ that minimizes the online cross-validated risk}
}

\newcommand{\argminD}{\arg\!\min}

\ShortHeadings{Personalized Online Machine Learning}{}
\firstpageno{1}

\begin{document}

\title{Personalized Online Machine Learning}

\author{\name Ivana Malenica* \email 
imalenica@berkeley.edu \\
       \addr Division of Biostatistics\\
       University of California, Berkeley
       \AND
       \name Rachael V. Phillips* \email 
       rachaelvphillips@berkeley.edu  \\
       \addr Division of Biostatistics\\
       University of California, Berkeley
       \AND
       \name Romain Pirracchio \email 
       romain.pirracchio@ucsf.edu \\
       \addr Department of Anesthesia and Perioperative Care \\
       University of California San Francisco
       \AND
       \name Antoine Chambaz \email 
       antoine.chambaz@parisdescartes.fr   \\
       \addr MAP5 (UMR CNRS 8145)\\
       Université de Paris, Paris, France
       \AND
       \name Alan Hubbard \email 
       hubbard@berkeley.edu   \\
       \addr Division of Biostatistics\\
       University of California, Berkeley
       \AND
       \name Mark van der Laan \email 
       laan@berkeley.edu \\
       \addr Division of Biostatistics\\
       University of California, Berkeley}

\maketitle

\begin{abstract}

In this work, we introduce the Personalized Online Super Learner (POSL) -- an online ensembling algorithm for streaming data whose optimization procedure accommodates varying degrees of personalization. Namely, POSL optimizes pre\-dic\-tions with respect to baseline covariates, so personalization can vary from completely individualized (i.e., optimization with respect to baseline covariate subject ID) to many individuals (i.e., optimization with respect to common baseline covariates). As an online algorithm, POSL learns in real-time. POSL can leverage a diversity of candidate algorithms, including online algorithms with different training and update times, fixed algorithms that are never updated during the procedure, pooled algorithms that learn from many individuals' time-series, and individualized al\-go\-rithms that learn from within a single time-series. POSL's ensembling of this hybrid of base learning strategies depends on the amount of data collected, the stationarity of the time-series, and the mutual characteristics of a group of time-series. In essence, POSL decides whether to learn across samples, through time, or both, based on the underlying (unknown) structure in the data. For a wide range of simulations that reflect realistic forecasting scenarios, and in a medical data application, we examine the performance of POSL relative to other current ensembling and online learning methods. We show that POSL is able to provide reliable predictions for time-series data and adjust to changing data-generating environments. We further cultivate POSL's practicality by extending it to settings where time-series enter/exit dynamically over chronological time.
\end{abstract}


\newpage

\section{Introduction}\label{sect1}

Predictive analytics with large data streams is a common task across many fields, including physics, medicine, engineering and finance. The insights drawn from these data typically come in the form of forecasts (predictions about the future), and inform subsequent action by the machine or user. The usefulness of the forecasts (interchangeably referred to as ``predictions'') often depends on their timeliness, accuracy, and uncertainty. For example, in the intensive care unit (ICU), it is imperative to quickly convert streams of data into predictions \citep{chan2020}. The same is true for high-frequency trading, where computers need to rapidly make beneficial decisions from their forecasts. Drawing from the COVID-19 pandemic, obtaining accurate forecasts in a timely manner is crucial for making evolving policy decisions \citep{altieri2020curating}. In these examples, and for real-world data streams in general, the observations are derived from dynamic environments, where time-series are ever-growing and evolve in possibly unforeseeable ways. 

In order for a machine to quickly adapt with the dynamic patterns in data streams, algorithmic strategies that regularly reassess the information learned from incoming data relative to historical data are crucial. The traditional machine learning and forecasting paradigm has been in the form of offline estimation, where a prediction algorithm (also referred to as ``learner'' and as shortened ``algorithm'') is updated with new batches of data by first adding the new data to all, or part, of the existing data, and then retraining the learner on the new training dataset. Because an offline algorithm's training data grows with each update, these strategies are generally not scalable when updates are frequent. Tools that assess the reliability of an algorithm, such as calibration diagnostics, can inform reactive updates for an offline algorithm. However, when new patterns are expected to emerge quickly and often in the time-series, real-time (as opposed to reactionary) learning is essential to maintain the reliability of the system. 

Online estimation, where an algorithm is updated with a new batch of data without revisiting past training data, has become a promising technique for learning from data streams in real-time. There is a growing body of literature on online algorithms and software, including online implementations of canonical time-series algorithms \citep{anava2013,hoi2018}. Some implementations are ensemble-based, combining forecasts from multiple algorithms as part of their procedure. Ensembling strategies have been shown to increase forecast accuracy; for instance, the most successful entries in the 2018 M4 Forecasting Competition were ensembling methods \citep{smyl2020, gilliland2020, shaub2020, pawlikowski2020}. \cite{hibon2005} showed empirically that the best combination of forecasts performed as well as the best individual forecast, and argued that choosing an individual algorithm from a set of available algorithms is more risky than choosing a combination. 

Many longstanding challenges exist in applying online learning strategies. Online learning strategies, just like all other machine learning, require data in order to perform well, and in some settings data accumulation over time is a luxury and is not guaranteed in practice. For instance, in order to forecast a patient's trajectory in the hospital, a purely online learning strategy would require following the patient for a long period of time beforehand, which is not practical for in-hospital forecasting applications. Another limitation of online learning is the phenomenon of ``catastrophic forgetting/interference'', in which the new information interferes with what the model has already learned. Catastrophic forgetting can result in sudden drops in performance and/or overwriting prior knowledge that could be informative again in the future \citep{Lee2020}. Constrained online learning/ensembling strategies offer the potential to reduce catastrophic forgetting events as the degree in which an algorithm is allowed to adjust its parameters at each update could be restricted. Also, online ensembling of a hybrid of offline algorithms and online algorithms  provides a means to address these limitations.

Considering that specification error tends to propagate in online learning settings, a principled methodology for algorithm selection and ensembling is warranted \citep{hibon2005, shaub2020}. However, despite the emerging popularity of online learning implementations and ensembling algorithms, literature at the intersection of these two fields is relatively scarce. Furthermore, a \textit{personalized} online ensembling paradigm that is grounded in statistical optimality theory, to fit and evaluate the performance among multiple diverse algorithms under an individualized optimization strategy, has only been described in the commercial/proprietary realm and has not yet been formally defined in the literature, to the best of our knowledge. This article proposes such a principled paradigm, and builds off developments presented in \cite{benkeser2018}.

\cite{benkeser2018} generalized an existing oracle result for independent and identically distributed (i.i.d.) data to time-series data, showing that the online algorithm exhibiting the best cross-validated performance is asymptotically equivalent to the oracle; the oracle selects the algorithm in the library (the set of candidate algorithms) that has the best performance with respect to the the true, unknown data-generating process (DGP) \citep{dudoit2003a,dudoit2003b, vaart2006, laan2006oracle, sl2007, benkeser2018, ecoto2021onestep}. As such, the oracle cannot be computed in practice, but it serves as a useful benchmark. These oracle results were further extended to the best combination of individual online algorithms, showing that the ensembling online learner's performance is asymptotically equivalent to an optimal ensemble that performs best with respect to the DGP \cite{benkeser2018}. \cite{benkeser2018} propose one form of cross-validation for identifying the best candidate online algorithm among a set of possible online learners, and applied this to training with batches of incoming data and with a single individual's data stream. 

In this work, we introduce a novel online ensembling algorithm --- Personalized Online Super Learner (POSL) --- that utilizes a diversity of time-series and ensembling methods, with the goal of optimizing baseline-covariate-level (including individual-level) forecasts. POSL leverages multiple candidate algorithms, training pooled (population based) and online individualized learners at each time step, while allowing for the ensembling to depend on the amount of data collected, status of stationarity, and residual noise. As such, POSL is not hindered by the limitations of purely online learning or purely offline learning as it draws on both strategies. Our proposed algorithm is able to adapt to the underlying structure in the data, allowing it to pick between relying on the structure through time (e.g., conditional stationarity as in \cite{benkeser2018}) or number of samples at each time point (when no structure in time is present, or not enough time points are collected). In addition, POSL allows for completely different data-generating distributions across time-series, with possibly only common baseline covariates --- thus, it is constructed to provide optimal forecasts for a specific unit, as opposed to a collection of time-series. Overall, POSL is able to provide accurate predictions for short time-series data and adjust to changing data-generating environments.

Our main framework consists of observing $n$ data structures over a finite number of time points, possibly from different data-generating distributions, where each observation is comprised of baseline and time-varying covariates and a response. We show that this setup is easily extended to settings where individual time-series enter/exit dynamically, with possibly different lengths and numbers of time-series observed at chronological time $t$. We propose several new online forms of cross-validation for single and multiple time-series, with varying sample dependence, all of which can be used to identify the best candidate algorithm in the current library. Building on theoretical foundations proposed by \cite{benkeser2018}, and under stronger assumptions than necessary for the setup we describe, we apply the online oracle inequalities to multiple time-series showing that the online algorithm with the best cross-validated performance is asymptotically equivalent with the performance of the oracle. Lastly, we propose an adaptive ensembling step, that bases the final ensemble weights on mutual characteristics of a group of time-series, allowing for a continuum of weight personalization that aims to increase predictive power of forecasts for shorter time-series.

\subsection{Outline}

 We formulate the general statistical estimation problem in Section~\ref{sect3}, including defining the target parameter and the loss-based paradigm for estimation. In Section~\ref{sect4}, we present several online cross-validation schemes, all of which are valid for POSL; proposed extensions to the multiple time-series settings can be found in Appendix~B. In Section~\ref{sect5}, we discuss how online cross-validation is used to identify the best performing individual online algorithm (discrete online super learner) and the best performing online ensemble of individual algorithms, and we define the oracle selector for POSL. In Section~\ref{sect6}, we extend the current formulation of POSL to adaptive enrollment, with possibly different length of time-series and number of available subjects at each time point. In Section~\ref{sect7} we conduct multiple simulation studies, and compare POSL to various ensembling and online methods currently available in the literature. In Section~\ref{sect8}, we provide a data analysis example for blood pressure forecasting. We conclude with a short discussion in Section~\ref{sect9}.
 
\section{Formulation of the Estimation Problem}\label{sect3}

In the following, we formalize the prediction task as an estimation problem, identifying the statistical target parameter as the minimizer of the risk induced by a valid, problem-specific loss function. 

\subsection{Data, Likelihood and the Statistical Model}\label{sect3.1}

We model a data structure under the shape of a random variable defined as $O_i = (O_i(t): t=1, \ldots, \gls{gls_tau})$, where \gls{gls_Ot} is a time $t$-specific $p$-dimensional variable for sample~$i$. We focus on situations where the time-varying part $O_i(t)$ decomposes as $O_i(t) = (W_i(t), Y_i(t))$, with \gls{gls_Wt} defining a vector of time-varying covariates occurring before the response variable \gls{gls_Yt}, both indexed by time $t$ and sample id $i$. We denote \gls{gls_X} as the vector of baseline covariates which, by definition, are initiated at $t=0$ and are not dependent on $t$. We view each $X_i$ and $O_i = (O_i(t) : t = 1, \ldots, \tau)$ as the sample $i$-specific baseline covariates and time-series, respectively. With that, we note that $X_i$ could simply be a function of $i$ itself, or continuous/discrete covariates which allow one to smooth across the subjects. We observe \gls{gls_n} independent realizations of random variables denoted as $(X_1,O_1), \ldots, (X_n,O_n)$. For convenience, we also introduce $X^n = (X_i : i=1, \ldots, n)$ and \gls{gls_On} $= (O_i : i = 1, \ldots, n)$ as the collections consisting of the $n$ units. Consequently, for each $i = 1, \ldots, n$, $O^n(t) = (O_i(t) : i = 1, \ldots, n)$ reflects the  collection of the $n$ $t$-specific $p$-dimensional variables observed at time $t$. 

We denote by \gls{gls_M} the statistical model; that is, the set of laws from which $(X^n,O^n)$ can be drawn. The more we know, or are willing to assume about the experiment that produces the data, the smaller is the model; this will be discussed momentarily. Let \gls{gls_P0} $\in \mathcal{M}$ be the true probability distribution of $(X^n,O^n)$. Moreover, let \gls{gls_Px} be the conditional distribution of $O_i$ given $X_i$ for each $i=1, \ldots, n$. When conditioning on $X_i$, we use the short notation $P_{0,X_i}$ instead of $P_{0,O_i|X_i}$ (not to be confused with $P_{0,X}$, the marginal distribution over the baseline covariates). We emphasize that $P_{0,X_i}$ could be just unit $i$ specific, as is the case when $X_i$ is simply a function of $i$ itself; alternatively, $P_{0,X_i}$ could be a smooth function of $X_i$, allowing one to smooth across the units. 
We let $p_0^n$ denote the density of $P_0^n$ with respect to (w.r.t) a measure $\mu^n$ that dominates all elements of $\mathcal{M}$. The joint likelihood of $(x^n,o^n)$ can be factorized according to the time-ordering as follows:
\begin{align}
\label{eq:model:calM:one}
p_0^n(x^n,o^n) &= \prod_{i=1}^{n} p_{0,x}(x_i) \prod_{t=1}^{\tau} p_{0,o_{i}(t)} (o_{i}(t) \ | \ x_i, \overline{o}_{i}(t-1)),
\end{align}
where $p_{0,x}$ marks the probability density for the baseline covariates, and $p_{0,o_{i}(t)}$ is the conditional density of $O_{i}(t)$ given $X_i$ and all the observed past until time $t$ for sample $i$. In particular, we define \gls{gls_Obar} as the $t$-specific history of the time-series for sample~$i$, with $\overline{O}_i(t-1) = (O_i(1),  \ldots, O_i(t-1))$ (note the convention $\overline{O}_i(0) = \emptyset$). 


In order to allow learning from a dependent process, we must make a few assumptions on the law of the data, $P_{0}^n$, through restrictions made on the statistical model~$\mathcal{M}$. In particular, we assume that each factor $p_{0,o_i(t)}(O_i(t)|X_i, \overline{O}_i(t-1))$ depends on the past through a fixed-dimensional summary measure \gls{gls_Z}. For some applications, the fixed dimensional summary measure $Z_i(t)$ covers a limited history, that is $Z_i(t) = (O_i(t-m), \ldots, O_i(t))$. In words, the dependent process has a finite memory $m$ allowing us to learn through time. Similarly, we could have defined $Z_i(t)$ to be a function of finite memory $m$ of a finite number of other time-series in unit $i$'s network. Another example of summary measures is  $Z_{i}(t) = t^{-1} \sum_{t'=1}^{t} O_i(t')$, with the means computed component-wise. The formulation of $Z_i$ is general enough to allow for different trends (e.g., seasonality), because the definition of $Z_i$ can involve $i$ itself.

Secondly, we define \gls{gls_Pcommon} as the \textit{common} conditional probability distribution of $O_i(t)$ given $Z_{i}(t-1)$ and $X_i$ under $P^n$. 
As we will describe later, this assumption is not crucial for the algorithm itself --- if not enough time points are collected, we rely on performance based on the number of trajectories. However, if online learning is to be useful, and achieve oracle results, some structure across time is necessary. We also stress that our formulation allows for $P_{0,O_i}$ to be a function of time $t$, making it possible for the proposed procedure to learn how much to rely on conditional stationarity over time. Thus, in light of \eqref{eq:model:calM:one} and the two above mentioned assumptions, the joint likelihood of $(x^n, o^n)$ under any element $P^n$ of the (constrained) statistical model $\mathcal{M}$ decomposes as
\begin{equation}
\label{eq:model:calM:two}
p^n(x^n,o^n) 
= \prod_{i=1}^{n} p_{x}(x_i) \prod_{t=1}^{\tau} p_{o_i} (o_{i}(t) \ | \ x_i, z_{i}(t-1)),
\end{equation}
where we extend the previously described notation with the substitution of $P^n$ ($p^n$) for $P_0^n$ ($p_0^n$). Note that $P_{0,O_i}$ is subject specific, and we don't assume a common across $i$ distribution; if all the time-series are drawn from the same distribution, we let the algorithm learn that. In the rest of the manuscript, we will deal explicitly with $P_{0,O_i}$ and $P_{0,X_i}$. 

The above derivation of \eqref{eq:model:calM:two} hinges on independence across subjects. While we write the likelihood as a product of both the number of samples ($n$) and time points ($t$), we emphasize that for deriving our main results, dependent on the asymptotics in time, we do not need to assume anything about dependence among subjects. Our statistical model $\mathcal{M}$ is, in essence, a model for a single time-series. With that, while independence across subjects allows us to have asymptotics in total number of time points observed across the $n$ subjects, dependence among subjects is also allowed under the described formulation. In particular, network dependence could be allowed simply by letting each  $Z_i(t-1)$ to summarize the whole past \gls{gls_Onbar} $= \{\overline{O}_1(t-1), \ldots,  \overline{O}_n(t-1)\}$ of $O^n$ at time $(t-1)$, or a subgroup specific past of its network, as opposed to the $i$-specific past $\overline{O}_i(t-1)$. 

\subsection{Statistical Target Parameter}\label{target}

Most prediction-based literature focuses on parameters of the population distribution $P_0^n$ or, as is the case for the time-series literature, on unit-specific forecasts. Our goal is not to understand the population distribution $P_0^n$. Instead, we focus on the parameters of the unit-specific conditional distribution $P_{0,X_i}$.

We define the relevant feature of the true data distribution we are interested in as the statistical target parameter. As in Section~\ref{sect3.1}, we assume that $P_{0}^n$ belongs to a statistical model $\mathcal{M}$, defined as a collection of possible common conditional distributions $P_{0,O_i}$ and marginal $P_{0,X}$ that could have given rise to the observed data. We define a parameter mapping, $\gls{gls_Psi} : \mathcal{M} \to \mathcal{D}$, from the model $\mathcal{M}$ into a space $\mathcal{D}$; and a parameter value, $\gls{gls_psi} \coloneqq \Psi(P)$ of $\Psi$ for a given $P \in \mathcal{M}$. The parameter space, corresponding to parameter mapping $\Psi$, is defined as $ \pmb{\Psi} \coloneqq \{\Psi(P) : P \in \mathcal{M} \} \subseteq \mathcal{D}$. 

In some cases, we might be interested in learning the entire conditional distribution $P_{0,X_i}$; however, frequently the actual goal is to learn a particular feature of the true distribution that satisfies a scientific question of interest. In particular, we are interested in forecasting -- hence, we define our estimand for the $i^{\text{th}}$ subject as:
\begin{equation}
    \Psi(P_{0}^n)(X_i, Z_i(t-1), t) = E_{P_{0,X_i}}[Y_i(t) | Z_i(t-1)], 
\end{equation}
where the expectation on the right hand side is taken w.r.t the conditional distribution $P_{0,X_i}$, and $\psi_0(X_i,Z_i(t-1),t) \coloneqq \gls{gls_estimand}$ is the prediction function evaluated at the truth for the $i^{\text{th}}$ subject at time $t$. In particular, we want to learn $(X,Z(t-1),t) \mapsto \Psi(P_0^n)(X,Z(t-1),t)$, where $Z(t-1)$ is fixed dimensional, and thereby obtain a prediction function for each unit $i$ that predicts $Y_i(t)$ with $\Psi(P)(X_i,Z_i(t-1),t)$ for $P \in \mathcal{M}$.

\subsection{Loss-based Parameter Definition and Estimation}\label{loss}

We define $L$ as a loss function; we emphasize that the chosen loss should be picked in accordance with the target parameter. Specifically, a valid loss function for a given parameter is defined as a function whose true conditional mean is minimized by the true value of the parameter. As such, let $L$ be a loss function adapted to the problem, i.e. a function that maps every $\Psi(P)$ to $L(\Psi(P)) : (x_i, y_i(t), z_i(t-1)) \mapsto L(\Psi(P))(x_i, y_i(t), z_i(t-1))$. With that, we define \gls{gls_loss} as a time $t$ and subject $i$ loss for $\Psi(P)$. Note that we could equivalently define $L$ a function that maps every $\psi$ to $L(\psi) : (x_i, y_i(t), z_i(t-1)) \mapsto L(\psi)(x_i, y_i(t), z_i(t-1))$ since $\psi \coloneqq \Psi(P)$. As our parameter of interest is a conditional mean, we could use the square error to define the loss; then we have that $L(\psi)(X_i,Y_i(t),Z_i(t-1)) = c(i,t)(Y_i(t) - \psi(X_i, Z_i(t-1), t))^2$, where $c(i,t)$ is the subject and time specific weight. Our accent on appropriate loss functions strives from their multiple use within our framework --- as a theoretical criterion for comparing the estimator and the truth, as well as a way to compare multiple estimators of the target parameter.

We define the true risk as the expected value of $L(\psi)(X_i,Y_i(t),Z_i(t-1))$ w.r.t the true conditional distribution $P_{0,O}$ across all individuals and times:
\begin{align}\label{rsk}
R(P_{0}^n, \psi) &= \frac{1}{n} \sum_{i=1}^n  \sum_{t=1}^{\tau} E_{P_{0,O_i}} [L(\psi)(X_i,Y_i(t),Z_i(t-1)) | X_i, Z_i(t-1)]\\
&= \frac{1}{n} \sum_{i=1}^n  \sum_{t=1}^{\tau} E_{P_{0,O_i}} [c(i,t)(Y_i(t) - \psi(X_i, Z_i(t-1),t))^2 | X_i,  Z_i(t-1)], \nonumber
\end{align}
where the second equality holds only when the loss function is valid for the target parameter; we simply illustrate what \gls{gls_risk} would be with squared error as a loss in \eqref{rsk}. The notation for true risk, $R(P_{0}^n, \psi)$, emphasizes that $\psi$ is evaluated w.r.t. the true data-generating distribution. Finally, we define $\psi_0$ as the minimizer over the true risk of all evaluated $\psi$ in the parameter space
\begin{equation}
\gls{gls_psi_0} = \argminD \limits_{\psi \in \pmb{\Psi}} R(P_{0}^n,\psi).
\end{equation}
The corresponding true risk is denoted as $\theta_0 = R(P_{0}^n,\psi_{0})$. 


The true risk establishes a true measure of performance for $\psi$, optimizing over all times. We note, however, that we could also define a true $i$-specific risk --- where the $i$-specific risk would measure the performance of $\psi$ for individual $i$ across all time points. Note that $\psi_0$ implies $\psi_{0,i}$ by evaluating at $X_i$, as $\psi_{0,i}$ is a prediction function given $X_i$. The $i$-specific expected loss measures the performance of the prediction function $\Psi$ for individual $i$ across all time points, optimizing the following equation:
\begin{equation}
\gls{gls_psi_0i} = \argminD \limits_{\psi \in \pmb{\Psi}} \sum_{t=1}^{\tau} E_{P_{0,O_i}} [L(\psi)(X_i,Y_i(t),Z_i(t-1)) | X_i, Z_i(t-1)],
\end{equation}
with optimal risk, corresponding to $\psi_{0,i}$, defined as $\theta_{0,i} = R(P_{0}^n,\psi_{0,i})$. 

The estimator mapping, \gls{gls_hatPsi}, is a function from the empirical distribution to the parameter space $\pmb{\Psi}$. Let \gls{gls_Pnt} denote the empirical distribution of $n$ time series collected until time $t$. In particular, $P_{n,t} \mapsto \hat{\Psi}(P_{n,t})$ represents a mapping from $P_{n,t}$, with $n$ time-series collected until time $t$, into a predictive function $\hat{\Psi}(P_{n,t})$. Further, the predictive function $\hat{\Psi}(P_{n,t})$ maps $(X_i,Z_i(t-1),t)$ into a time- and subject-specific outcome, $Y_{i}(t)$. We emphasize that $\hat{\Psi}(P_{n,t})$ can map any $(X_i, Z_i(s-1),s)$ into a time $s$ prediction, even for $s > t$ under stationarity conditions. We can write $\psi_{n,t}(X_i,Z_i(t-1),t) \coloneqq \hat{\Psi}(P_{n,t})(X_i,Z_i(t-1),t)$ as the predicted outcome for unit $i$ of the estimator $\hat{\Psi}(P_{n,t})$ at time $t$, based on $(X_i,Z_i(t-1),t)$. We define the conditional risk as the risk for \gls{gls_psi_nt} with respect to the true, unknown data-generating distribution~$P_{0}^n$, denoted as $\tilde{\theta}_n = R(P_{0}^n, \psi_{n,t})$. The naive risk is defined as $\hat{\theta}_n = R(P_{n,t}, \psi_{n,t})$. In order to obtain an unbiased estimate of the true conditional risk, we resort to appropriate cross-validation for dependent data, as described in the next section. 


\section{Cross-validation for Dependent Data}\label{sect4}
 
Let \gls{gls_C} denote, at minimum, the time $s$- and unit $i$-specific record $C(i,s,\cdot) = (X_i, Z_i(s-1), Y_i(s), \cdot)$. The general formulation of $C(i,s,\cdot)$ allows us to add identifying information (in addition to time and sample id) needed to construct valid cross-validation scheme; for instance, for dynamic enrollment/exit dates, $C(i,s,\cdot)$ might include enrollment and exit time for a time-series as well. If no additional information is included, we write $C(i,s,\cdot) = C(i,s)$. To derive a general representation for cross-validation (CV), we also define a time $t$ specific split vector $B_t$, where $t$ indicates the final time-point of the currently available data where for all $1 \leq i \leq n$, $B_t(i, \cdot) \in \{-1,0,1\}^t$. Let \gls{gls_v} be a particular $v$-fold, where $v$ range from $1$ to $V$. A realization of $B_t$ defines a particular split of the learning set into corresponding three disjoint subsets,
\[
  B_t^v(i,s,\cdot)=\begin{cases}
               -1, \ \  C(i,s,\cdot) \  \text{not used}\\
                \ \ \  0, \ \ C(i,s,\cdot) \  \text{in the training set}\\
                 \ \ \ 1, \ \ C(i,s,\cdot) \  \text{in the validation set,}
            \end{cases}
\]
where \gls{gls_Bn} reflects the $v$-fold assignment of, at minimum, unit $i$ at time point $s$ for split $B_t^v$ trained on data until time $t$. For each $t$, let \gls{gls_Pnt_0} denote the empirical distribution of the training sample until time $t$. Similarly, we define \gls{gls_Pnt_1} as the empirical distribution of the validation set. Let $\gls{gls_nt_0} = \sum_{v=1}^V \sum_{i=1}^n \sum_{s=1}^{t} \mathbb{I}(B_t^v(s,i,\cdot)=0)$ and $\gls{gls_nt_1} = \sum_{v=1}^V \sum_{i=1}^n \sum_{s=1}^{t} \mathbb{I}(B_t^v(s,i,\cdot)=1)$ denote the number of observations in the training and validation sets for split $B_t$, respectively, over all folds $v$ until time $t$. We let \gls{gls_Bt_0} denote all the $(i,s,\cdot)$ indexes in the training set, and \gls{gls_Bt_1} all indexes in the validation set for fold $v$. In general, we use different time-series cross-validation schemes to evaluate how well an estimator trained on specific samples' past data is able to predict an outcome for specific samples in the future. We now give relevant cross-validation schemes that are supported by the theoretical results for our proposed algorithm.



\subsection{Rolling Origin cross-validation}\label{origin}

Rolling origin cross-validation scheme lends itself to online cross-validation-based ensemble learning, as described by \cite{benkeser2018}. In general, the rolling origin scheme defines an initial training set and, with each iteration, the size of the training set grows by $m$ observations until we reach time $t$ for split $B_t$. Whether or not the samples in the training set are also present in the validation set is optional, but classically, rolling origin cross-validation represents the scenario where the training and validation points are evaluated on the same time-series. Regardless of which samples are included in the training and validation sets, time points included in the training set always occur before the validation set time points; in addition, there might be a gap between training and validation times of size $h$. We define cross-validation folds $v = 1, \ldots, V$ with $B_t^v$ for a single unit as follows:
\[
  B_t^v (i,s,\cdot) = \begin{cases}
               \ -1, \ \ C(i,s,\cdot) \  \text{not used, }\\
               \hspace{1.5cm} s \in \{n_{t,v_1}^{0} + m \times (v-1) + 1 : n_{t,v_1}^{0} + m\times (v-1) + h\} \\
                \ \ \  0, \ \ C(i,s,\cdot) \  \text{in the training set, } \\ 
                \hspace{1.5cm} s \in \{1 : n_{t,v_1}^{0} + m \times (v-1)\} \\
                 \ \ \ 1, \ \ C(i,s,\cdot) \  \text{in the validation set, } \\
                 \hspace{1.5cm} s \in \ \{n_{t,v_1}^{0} + m \times (v-1) + h +1 \ : \ n_{t,v}^{0} + m \times (v-1) + h + n_{t,v}^{1}\}
            \end{cases}
\]
where $n_{t,v_1}^{0}$ is the size of the training set at $v=1$, and $n_{t,v}^{1}$ is the size of the validation set for all $v$. An example of rolling origin cross-validation is illustrated in Figure \ref{fig::ro_cv} for $V=3$.


\begin{figure}[h!]
\centering
\begin{flushleft} \text{$v$ = 1} \end{flushleft}
\begin{tikzpicture}
\draw[thick] (0,0) -- (14,0) node[font=\scriptsize, below left=10pt]{};

\foreach \x in {1,...,12} \draw (\x cm,2pt) -- (\x cm,-2pt);

\foreach \x/\descr in {1/0, 2/5, 3/10, 4/15,  5/20, 6/25, 7/30, 8/35,
  9/40, 10/45, 11/50, 12/55} \node[font=\scriptsize, text height=1.75ex,
text depth=.5ex] at (\x,-.3) {$\descr$};

\draw [thick] (1,0.7)  -- ++(3,0)
node [black,midway,above=4pt, font=\scriptsize] {Training Set}; \draw
[thick] (5,0.7) -- ++(2,0) node
[black,midway,above=4pt, font=\scriptsize] {Validation Set}; \draw
[thick] (9,0.7) -- ++(0,0) node
[black,midway,above=4pt, font=\scriptsize] {};

\foreach \x/\perccol in {1/25,2/25,3/25} \draw[lightgray!\perccol!red, line
width=4pt] (\x,.5) -- +(1,0);

\foreach \x/\perccol in {5/25,6/25} \draw[lightgray!\perccol!blue, line width=4pt](\x,.5) -- +(1,0);

\end{tikzpicture}

\begin{flushleft} \text{$v$ = 2} \end{flushleft}
\begin{tikzpicture}
\draw[thick] (0,0) -- (14,0) node[font=\scriptsize, below left=10pt]{};

\foreach \x in {1,...,12} \draw (\x cm,2pt) -- (\x cm,-2pt);

\foreach \x/\descr in {1/0, 2/5, 3/10, 4/15,  5/20, 6/25, 7/30, 8/35,
  9/40, 10/45, 11/50, 12/55} \node[font=\scriptsize, text height=1.75ex,
text depth=.5ex] at (\x,-.3) {$\descr$};

\draw [thick] (1,0.7)  -- ++(5,0)
node [black,midway,above=4pt, font=\scriptsize] {Training Set}; \draw
[thick] (7,0.7) -- ++(2,0) node
[black,midway,above=4pt, font=\scriptsize] {Validation Set}; \draw
[thick] (11,0.7) -- ++(0,0) node
[black,midway,above=4pt, font=\scriptsize] {};

\foreach \x/\perccol in {1/25,2/25,3/25,4/25,5/25} \draw[lightgray!\perccol!red, line
width=4pt] (\x,.5) -- +(1,0);

\foreach \x/\perccol in {7/25,8/25} \draw[lightgray!\perccol!blue, line width=4pt](\x,.5) -- +(1,0);
\end{tikzpicture}

\begin{flushleft} \text{$v$ = 3} \end{flushleft}
\begin{tikzpicture}

\draw[thick] (0,0) -- (14,0) node[font=\scriptsize, below left=10pt]{time (s) };

\foreach \x in {1,...,12} \draw (\x cm,2pt) -- (\x cm,-2pt);

\foreach \x/\descr in {1/0, 2/5, 3/10, 4/15,  5/20, 6/25, 7/30, 8/35,
  9/40, 10/45, 11/50, 12/55} \node[font=\scriptsize, text height=1.75ex,
text depth=.5ex] at (\x,-.3) {$\descr$};

\draw [thick] (1,0.7)  -- ++(7,0)
node [black,midway,above=4pt, font=\scriptsize] {Training Set}; \draw
[thick] (9,0.7) -- ++(2,0) node
[black,midway,above=4pt, font=\scriptsize] {Validation Set}; \draw
[thick] (11,0.7) -- ++(0,0) node
[black,midway,above=4pt, font=\scriptsize] {};

\foreach \x/\perccol in {1/25,2/25,3/25,4/25,5/25,6/25,7/25} \draw[lightgray!\perccol!red, line
width=4pt] (\x,.5) -- +(1,0);

\foreach \x/\perccol in {9/25,10/25} \draw[lightgray!\perccol!blue, line width=4pt](\x,.5) -- +(1,0);
\end{tikzpicture}

\caption{Rolling origin cross-validation scheme for $V=3$ folds with first window size $n_{t,v_1}^0=15$, validation size $n_{t,v}^1=10$, batch size $m=10$ and gap $h=5$ for sample~$i$.}
\label{fig::ro_cv}
\end{figure}

A variant of rolling origin scheme which accounts for sample dependence is the rolling-origin-$V$-fold cross-validation. In contrast to the canonical rolling origin CV, samples in the training and validation set are not the same, as this variant of the rolling origin CV encompasses of $V$-fold CV in addition to the time-series setup. The predictions are evaluated on the future times of time-series units not seen during the training step, allowing for dependence in both samples and time. In order to characterize $B_t^v$, we add sample id in addition to $n_{t,v_1}^0$, $n_{t,v}^1$, $m$ and $h$ in Figure \ref{fig::rov_cv}. We show an example with $V=2$ $V$-folds (i.e., splitting across samples, which we denote as v') and $V=2$ time-series CV folds (denoted as v) in Appendix B, Figure \ref{fig::rov_cv}.

\subsection{Rolling Window Cross-validation}\label{window}

Instead of adding more time points to the training set per each iteration, the rolling window cross-validation scheme ``rolls'' the training sample forward by $m$ time units. The rolling window scheme might be considered in parametric settings when one wishes to guard against moment or parameter drift that is difficult to model explicitly; it is also more efficient for computationally demanding settings such as streaming data, in which large amounts of training data cannot be stored. In contrast to rolling origin CV, the training sample size for each iteration of the rolling window scheme is always $n_{t,v}^0$ for all $v$. We emphasize that the rolling window CV could also be viewed as a subset of the rolling origin CV where only recent data are used for training; with that, we can incorporate a variant of the rolling window approach by using a learner that only trains on the recent past as one of the candidates in a library. In such a scenario, a rolling origin CV could be used to evaluate the final loss, but we might incorporate learners that train on fewer training time-points (as opposed to all training points from the start) thus adjusting to changes over time. In all the further sections and theoretical results, we consider rolling window as a subset of a rich class consisting of the rolling origin CV. Finally, we define cross-validation folds $v = 1, \ldots, V$ with $B_t^v$ and gap of size $h$ for a single time-series as shown below. We illustrate canonical rolling window cross-validation in Figure \ref{fig::rw_cv}, and the rolling-window-$V$-fold cross-validation which accounts for sample dependence in Figure \ref{fig::rwv_cv} (allocated to Appendix B).

\[
  B_t^v(i,s,\cdot)=\begin{cases}
               \ -1, \ \ C(i,s,\cdot) \  \text{not used,} \\
               \hspace{1.5cm} s \in \ \{n_{t,v}^0 + m\times (v-1) +1 : n_{t,v}^0 + m\times (v-1) + h\} \\
                \ \ \  0, \ \ C(i,s,\cdot) \  \text{in the training set, } \\ 
                \hspace{1.5cm} s \in \ \{n_{t,v}^0 + m\times (v-1) - n_{t,v}^0 : n_{t,v}^0 + m\times (v-1)\} \\
                 \ \ \ 1, \ \ C(i,s,\cdot) \  \text{in the validation set, } \\
                 \hspace{1.5cm} s \in \ \{n_{t,v}^0 + m\times (v-1) + h + 1 \ : \ n_{t,v}^0 + m\times (v-1) + h + n_{t,v}^1\}.
            \end{cases}
\]


\begin{figure}[H]
\centering
\begin{flushleft} \text{$v$ = 1} \end{flushleft}
\begin{tikzpicture}
\draw[thick] (0,0) -- (14,0) node[font=\scriptsize, below left=10pt]{};

\foreach \x in {1,...,12} \draw (\x cm,2pt) -- (\x cm,-2pt);

\foreach \x/\descr in {1/0, 2/5, 3/10, 4/15,  5/20, 6/25, 7/30, 8/35,
  9/40, 10/45, 11/50, 12/55} \node[font=\scriptsize, text height=1.75ex,
text depth=.5ex] at (\x,-.3) {$\descr$};

\draw [thick] (1,0.7)  -- ++(3,0)
node [black,midway,above=4pt, font=\scriptsize] {Training Set}; \draw
[thick] (5,0.7) -- ++(2,0) node
[black,midway,above=4pt, font=\scriptsize] {Validation Set}; \draw
[thick] (9,0.7) -- ++(0,0) node
[black,midway,above=4pt, font=\scriptsize] {};

\foreach \x/\perccol in {1/25,2/25,3/25} \draw[lightgray!\perccol!red, line
width=4pt] (\x,.5) -- +(1,0);

\foreach \x/\perccol in {5/25,6/25} \draw[lightgray!\perccol!blue, line width=4pt](\x,.5) -- +(1,0);

\end{tikzpicture}

\begin{flushleft} \text{$v$ = 2} \end{flushleft}
\begin{tikzpicture}

\draw[thick] (0,0) -- (14,0) node[font=\scriptsize, below left=10pt]{};

\foreach \x in {1,...,12} \draw (\x cm,2pt) -- (\x cm,-2pt);

\foreach \x/\descr in {1/0, 2/5, 3/10, 4/15,  5/20, 6/25, 7/30, 8/35,
  9/40, 10/45, 11/50, 12/55} \node[font=\scriptsize, text height=1.75ex,
text depth=.5ex] at (\x,-.3) {$\descr$};

\draw [thick] (3,0.7)  -- ++(3,0)
node [black,midway,above=4pt, font=\scriptsize] {Training Set}; \draw
[thick] (7,0.7) -- ++(2,0) node
[black,midway,above=4pt, font=\scriptsize] {Validation Set}; \draw
[thick] (11,0.7) -- ++(0,0) node
[black,midway,above=4pt, font=\scriptsize] {};

\foreach \x/\perccol in {3/25,4/25,5/25} \draw[lightgray!\perccol!red, line
width=4pt] (\x,.5) -- +(1,0);

\foreach \x/\perccol in {7/25,8/25} \draw[lightgray!\perccol!blue, line width=4pt](\x,.5) -- +(1,0);
\end{tikzpicture}

\begin{flushleft} \text{$v$ = 3} \end{flushleft}
\begin{tikzpicture}

\draw[thick] (0,0) -- (14,0) node[font=\scriptsize, below left=10pt]{time (s)};

\foreach \x in {1,...,12} \draw (\x cm,2pt) -- (\x cm,-2pt);

\foreach \x/\descr in {1/0, 2/5, 3/10, 4/15,  5/20, 6/25, 7/30, 8/35,
  9/40, 10/45, 11/50, 12/55} \node[font=\scriptsize, text height=1.75ex,
text depth=.5ex] at (\x,-.3) {$\descr$};

\draw [thick] (5,0.7)  -- ++(3,0)
node [black,midway,above=4pt, font=\scriptsize] {Training Set}; \draw
[thick] (9,0.7) -- ++(2,0) node
[black,midway,above=4pt, font=\scriptsize] {Validation Set}; \draw
[thick] (13,0.7) -- ++(0,0) node
[black,midway,above=4pt, font=\scriptsize] {};

\foreach \x/\perccol in {5/25,6/25,7/25} \draw[lightgray!\perccol!red, line
width=4pt] (\x,.5) -- +(1,0);

\foreach \x/\perccol in {9/25,10/25} \draw[lightgray!\perccol!blue, line width=4pt](\x,.5) -- +(1,0);
\end{tikzpicture}

\caption{Rolling window cross-validation scheme for $V=3$ time-series folds with window size $n_{t,v}^0=15$, horizon $n_{t,v}^1=10$, batch $m=10$ and gap $h=5$ for sample $i$.}
\label{fig::rw_cv}
\end{figure}

\section{Personalized Online Super Learner}\label{sect5}

\subsection{Online Cross-validation Selector}\label{cv_selector}

Suppose we have $K$ candidate estimators, \gls{gls_hatPsi_k}, and recall the definition of an estimator from subsection \ref{loss} (page \pageref{loss}). In order to evaluate performance of each $\hat{\Psi}_k$, we use cross-validation for dependent data to estimate the average loss for each candidate. In particular, each $\hat{\Psi}_k$ is trained on the training set until time $t$, using $P_{n,t}^0$ and resulting in a predictive function $\gls{gls_psi_ntk_0} \coloneqq \hat{\Psi}_k(P_{n,t}^0)$ for $k= 1, \ldots, K$. We define the online cross-validated risk for each candidate estimator as:
\begin{align}
\gls{gls_risk_cv} &= \sum_{j=1}^{t} \sum_{v=1}^V \sum_{(i,s) \in \mathcal{B}_{j,v}^1} L(\hat{\Psi}_{k}(P_{n,j}^0))(X_i,Y_i(s), Z_i(s-1)) \\
&= \sum_{j=1}^{t} \sum_{v=1}^V \sum_{(i,s) \in \mathcal{B}_{j,v}^1} L(\psi_{n,j,k}^0)(C(i,s)), \nonumber
\end{align}
where $R_{CV}(P_{n,t}^1, \hat{\Psi}_k(\cdot))$ is the cumulative performance of $\hat{\Psi}_k$ trained on training sets and evaluated on corresponding validation samples across all time points until time~$t$. For instance, while $\hat{\Psi}_k(P_{n,t}^0)$ is trained on the training set $P_{n,t}^0$, its performance will be over the validation set $P_{n,t}^1$. Additionally, if $\hat{\Psi}_k$ is an online estimator, then the online cross-validated risk is also an online estimator. For the squared error loss mentioned in Section \ref{loss} where $c(i,j) = 1$ (and is thus omitted), we rewrite the above online cross-validated risk as:
\begin{align}
R_{CV}(P_{n,t}^1, \hat{\Psi}_k(\cdot)) &= \sum_{j=1}^{t} \sum_{v=1}^V \sum_{(i,s) \in \mathcal{B}_{j,v}^1} (Y_i(s) - \hat{\Psi}_k(P_{n,j}^0)(X_i, Z_i(s-1), s))^2.
\end{align}
The online cross-validated risk estimates the following true online cross-validated risk, denoted as $R_{CV}(P_{0}^n, \hat{\Psi}_k(\cdot))$ and expressed as
\begin{align} 
\gls{gls_risk_cv_true} &= \sum_{j=1}^{t} \sum_{v=1}^V \sum_{(i,s) \in \mathcal{B}_{j,v}^1} E_{P_{0,O}}[L(\psi_{n,j,k}^0)(C(i,s)) | X_i, Z_i(s-1)].
\end{align}
Note that $R_{CV}(P_{0}^n, \hat{\Psi}_k(\cdot))$ reflects the true average loss for the candidate estimator with respect to the true conditional distribution $P_{0,O_i}$.
As opposed to the true online cross-validated risk, $R_{CV}(P_{n,t}^1, \hat{\Psi}_k(\cdot))$ gives an empirical measure of performance for each candidate estimator $k$ trained on training data until time $t$. In light of that, we define the discrete online cross-validation selector as:
\begin{align}
\gls{gls_knt} = \argminD_{k=1,\ldots, K} R_{CV}(P_{n,t}^1, \hat{\Psi}_k(\cdot)),
\end{align}
which is the estimator that minimizes the online cross-validated risk. 
The discrete (online) super learner is the estimator that at each time point $t$ uses the estimates from the discrete online cross-validation selector --- for time $t$, we have $\gls{gls_psi_discrete_0} \coloneqq \hat{\Psi}_{k_{n,t}}(P_{n,t}^0)$. We emphasize that the discrete super learner can switch from one learner to another as $t$ progresses, in response to accumulating more data and detecting changes in the time-series. Finally, if all the candidate estimators are online estimators, the discrete (online) super learner is itself an online estimator. 

\subsection{Defining the Gold Standard Oracle Selector}\label{oracle_selector}

In order to study performance of an estimator of $\psi_{0}$, we construct loss-based dissimilarity measures. First, we define $i$-specific loss-based dissimilarities for the $k^{\text{th}}$ estimator, $\hat{\Psi}_k$, trained until time $t$ as
\begin{equation}
\gls{gls_d0_i} = \sum_{j=1}^{t} \sum_{v=1}^V \sum_{(s) \in \mathcal{B}_{j,v}^1} E_{P_{0,O_i}}\bigg[ \Big( L(\psi_{n,j,k,i}) - L(\psi_{0,i})\Big) (C(i,s)) \bigg| X_i, Z_i(s-1)\bigg],
\end{equation}
which compares performance of the cross-validated estimator to the true parameter. Note that the training and validation sample is just sample $i$. We further define the measure $\gls{gls_d0}= \frac{1}{n} \sum_{i=1}^n d_{0,t}(\psi_{n,t,k,i}, \psi_{0,i})$ as an average of $i$-specific loss-based dissimilarities over all the samples until time $t$ for the $k^{\text{th}}$ estimator; 
$d_{0,t}(\psi_{n,t,k}, \psi_{0})$ reflects how far $\psi_{n,t,k}$ is from $\psi_{0}$ over all available times and samples in terms of the chosen loss. We define the time $t$ \textit{oracle} selector as the unknown estimator that uses the candidate closest to the truth in terms of the defined dissimilarity measure:
\begin{align}
\gls{gls_knt_oracle} = \argminD_{k=1,\ldots,K} d_{0,t}(\psi_{n,t,k}, \psi_{0}).
\end{align}
Due to it being a function of the true conditional mean, the oracle selector cannot be computed in practice. However, we can utilize it as benchmark in order to describe performance of the online cross-validation based estimator. In the Appendix A Theorem 1, assuming conditional stationarity (as proposed in Section \ref{sect3.1}, assumption not necessary for the algorithm function), 
we appropriate the work from \cite{benkeser2018} to multiple time-series using the cross-validation schemes described in Section \ref{sect3}. In particular, Appendix A Theorem 1 shows that the performance of the discrete POSL is asymptotically equivalent to that of the oracle selector. The result relies on the martingale finite-sample inequality by \cite{handel2009} to show that, as $t \rightarrow \infty$, 
\begin{align}
\frac{d_{0,t}(\psi_{n,t,k_{n,t}}, \psi_{0})}{d_{0,t}(\psi_{n,t,\overline{k}_{n,t}}, \psi_{0})} \rightarrow_p 1, 
\end{align}
under conditional stationarity and additional conditions specified in Appendix A.

\subsection{Ensemble of Candidate Estimators}\label{ensemble}

In this section, we consider a more flexible online learner that generates a weighted combination of candidate estimators. Let \gls{gls_hatPsi_alpha} be a function of empirical distribution ($P_{n,t}$, at any $t$) generating an ensemble of $K$ estimators $(\hat{\Psi}_1, \ldots, \hat{\Psi}_K)$ indexed by a vector of coefficients $\alpha$, where $\alpha = (\alpha_1, \ldots, \alpha_K)$. For example, $\hat{\Psi}_{\alpha}$ could represent a convex linear combination:
\begin{equation*}
    \hat{\Psi}_{\alpha} = \sum_{k=1}^K \alpha_k \hat{\Psi}_{k},
\end{equation*}
such that $\sum_{k=1}^K \alpha_k = 1$ and for all $\alpha_k$, $\alpha_k \geq 0$. 
We define conditional meta-learning by allowing the weight vector to depend on the baseline covariates $X$, where $\alpha(X) = \{\alpha_1(X), \ldots, \alpha_K(X)\}$ with $\sum_{k=1}^K \alpha_k(X) = 1$ and for all $\alpha_k(X)$, $\alpha_k(X) \geq 0$. For example, we can define $\alpha(X)$ by considering a parametric family $\mathcal{H} = \{\alpha_{\beta}:\beta\in\mathbb{B}\}$ where
\begin{equation*}
   \alpha_{\beta}(X) = \frac{\exp{(\beta_{k,1} + \beta_{k,2}X)}}{\sum_{k=1}^K \exp{(\beta_{k,1} + \beta_{k,2}X)}}.
\end{equation*}
To alleviate notation, we define $\alpha$ as an universal vector of coefficients (including conditional metalearning) in further sections.
Let $\hat{\Psi}_{\alpha} = \sum_{k=1}^K \alpha_k \hat{\Psi}_{k}$, so that the predictive function based on the training set $P_{n,t}^0$ is given by $\gls{gls_psi_nta_0} \coloneqq \sum_{k=1}^K \alpha_k \hat{\Psi}_{k}(P_{n,t}^0)$ with $\alpha \in \mathcal{H}$. We define a $\mathcal{H}$-specific online cross-validation selector for the ensemble as:
\begin{align}
\gls{gls_alpha} &= \argminD_{\alpha \in \mathcal{H}} R_{CV}(P_{n,t}^1, \hat{\Psi}_{\alpha}(\cdot)) \\
&= \argminD_{\alpha \in \mathcal{H}} \sum_{j=1}^{t} \sum_{v=1}^V \sum_{(i,s) \in \mathcal{B}_{j,v}^1}
(Y_i(s) - \hat{\Psi}_{\alpha}(P_{n,j}^0)(X_i, Z_i(s-1), s))^2, \nonumber
\end{align} 
where the loss is defined as the mean squared error. We can define an oracle selector for this class of estimators as the choice of weights that minimizes the true average of the loss-based dissimilarity:
\begin{align}
\gls{gls_alpha_oracle} &= \argminD_{\alpha \in \mathcal{H}} d_{0,t}(\psi_{n,t,\alpha}^0, \psi_{0}) \\
&= \argminD_{\alpha \in \mathcal{H}} \sum_{j=1}^{t} \sum_{v=1}^V \sum_{(i,s) \in \mathcal{B}_{j,v}^1} E_{P_{0,O_i}}\bigg[ \Big( L(\psi_{n,j,\alpha}^0) - L(\psi_{0})\Big) (C(i,s)) \bigg| X_i, Z_i(s-1)\bigg] . \nonumber
\end{align}
The results from Theorem 1 in Appendix A extend to all meta-learning, as the performance of the online cross-validated ensemble is asymptotically equivalent to the oracle ensemble of candidate estimators as $t$ goes to infinity, under conditional stationarity and conditions outlined in Appendix A,
\begin{align}
\frac{d_{0,t}(\psi_{n,t,\alpha_{n,t}}, \psi_{0})}{d_{0,t}(\psi_{n,t,\overline{\alpha}_{n,t}}, \psi_{0})} \rightarrow_p 1.
\end{align}

We note that one could also define a sequence of $\mathcal{H}_m$-specific online Super Learners, ranging from highly parametric to nonparametric for $m = 1, \ldots, M$, possibly stratified by the subject itself. Then, the online ensemble would have candidate algorithms $\hat{\Psi}_k$ for $k=1, \ldots, K$ augmented with a collection of online Super Learners indexed by weight classes $\mathcal{H}_m$, $m = 1, \ldots, M$. In this matter, the discrete online Super Learner would adaptively determine the optimal level of data adaptivity of the meta-learner, based on many candidates for the online discrete Super Learner considered. For example, depending on the number of subjects $n$ and time $t$, the choice $k_{n,t}$ of the online cross-validation selector might switch from discrete online Super Learner based on $K$ algorithms to more aggressive online Super Learner indexed by a more flexible weight class over time. 

\subsection{Practical Considerations for the Personalized Online Super Learner}\label{practical}

Due to the continuously updating procedure that allows the algorithm to evolve over time, POSL generalizes to a diversity of data streams. POSL accomodates varying degrees of personalization, such as within-covariate or within-subject; and it can handle multiple (potentially time-varying) dependence structures, from individual time series to networks of connected individuals. We delineate one version of POSL in Algorithm \ref{posl}, which can train online or in batches, and benefits from learning both from other subjects and from the history of the target individual's trajectory. We define Historical learners as $K_H$ learners generating a pooled (across individuals and time before $t$)  estimator $\hat{\Psi}_k(P_{n,t}^0)$ for algorithm $k \in K_H$, trained on samples $j = 1, \ldots n$ in the training sample. The motivation behind Historical learners is to provide an initial estimate based on previously collected trajectories (or multiple concurrently collected time-series), convenient for forecasting early in the trajectory for individual $i$; here, we don't need conditional stationarity for POSL to do well. Historical learners can be trained on time-series data collected even before the trajectory of interest is sampled, and can be updated at specific time points $t_s$ based on the computational efficiency. We note that Historical learners can generate a historical online Super Learner as well, which thus provides another candidate online estimator. Similarly, we define the Individual learners as $K_I$ learners applied to $P_{n,t}$, which stratify per subject when training. With that, we generate an individual estimator $\hat{\Psi}_{S,k}(P_{i,t}^0)$ for algorithm $k \in K_I$, individualized to sample target~$i$. Individual learners train on the training data by stratify by id, and predict the outcome in the future according to the forecast horizon. The $K_I$ candidate learners are possibly time-series learners, and are frequently updated in an online or batch fashion in order to accommodate the continuously incoming data. As for historical learners, individual learners can also form an individual online Super Learner, which becomes another candidate in the POSL library. With this formulation, we allow the POSL to leverage cross-validation in order to choose between pooled and individual fits at each time point $t$ --- in essence, allowing the algorithm to choose an appropriate structure from the data (learning from samples if no conditional stationarity is present or learning through time). This results in a natural adaptation to the amount of available data and the stationarity of the individual target time-series. The final discrete and full online super learner for the POSL is generated based on all the samples until specified time $t$. The cross-validation selector $k_{{n},t}$ reflects an optimized ensemble among all the available learners; the candidate learners reflect a collection of online algorithms which range in how much they use the current time series (stratify by id) and use the historical data (pool across all available time-series). All simulations in Section \ref{sect7} test the version of the POSL described in Algorithm \ref{posl}.

\begin{algorithm}[H] \caption{Personalized Online Super Learner}\label{posl}
\begin{algorithmic}[1] 
\State $t_s$: time steps at which Historical learner fit is updated.
\State $K_{H}$: Historical candidate learners.
\State $K_{I}$: Individual candidate learners.
\State $K$: All candidate learners, $K_{H} \cup K_{I}$.
\State $k$: Any learner among candidate learners.
\State 
\Procedure{Historical Learner}{n,t} 
\State
    Return $\hat{\Psi}(P_{n,t}^0)$, trained using using all available units ($i = 1, \ldots, n$) for any $t$.
    \State 
    \EndProcedure
    \Procedure{Individual Learner}{i,t} 
    \State
    Return $\hat{\Psi}_S(P_{n,t}^0)$, for any $t$ where we stratify by sample $i$.
    \State 
    \EndProcedure
    
\While{$t < \tau$}
\For{$k \in K_{H}$}
    \If{$t \in t_s$}
    \State Run \textsc{Historical Learner}(n,t), return $\psi_{n,t,k}^H = \hat{\Psi}_k(P_{n,t}^0)$.
    \Else{}
    \State $\psi_{n,t,k}^H = \hat{\Psi}_k(P_{n,t_{s-1}}^0)$.
    \EndIf
    \EndFor
\For{$k \in K_{I}$}
    \State Run \textsc{Individual Learner}(i,t), return $\psi_{i,t,k}^I = \hat{\Psi}_{S,k}(P_{n,t}^0)$.
    \EndFor
    
\If{\textsc{Discrete Online Super Learner}}
\State Return $\psi_{{n},t,k_{{n},t}}$, where $k_{{n},t} = \argminD_{k} R_{CV}(P_{{n},t}, \{\psi_{n,t,k}^H, \psi_{i,t,k}^I\})$. 
 \EndIf
 
\If{\textsc{Ensemble Online Super Learner}}
\State Return $\psi_{{n},t,\alpha_{n,t}}$, where $\alpha_{n,t} = \argminD_{\alpha} d_{0,t}(\{\psi_{n,t,k}^H, \psi_{i,t,k}^I\}, \psi_{0})$.
\EndIf
\EndWhile
\end{algorithmic}
\end{algorithm}


\section{Personalized Online Super Learner with Dynamic Streams}\label{sect6}

\begin{figure}
 \makebox[\textwidth][c]{\includegraphics[width=1\textwidth]{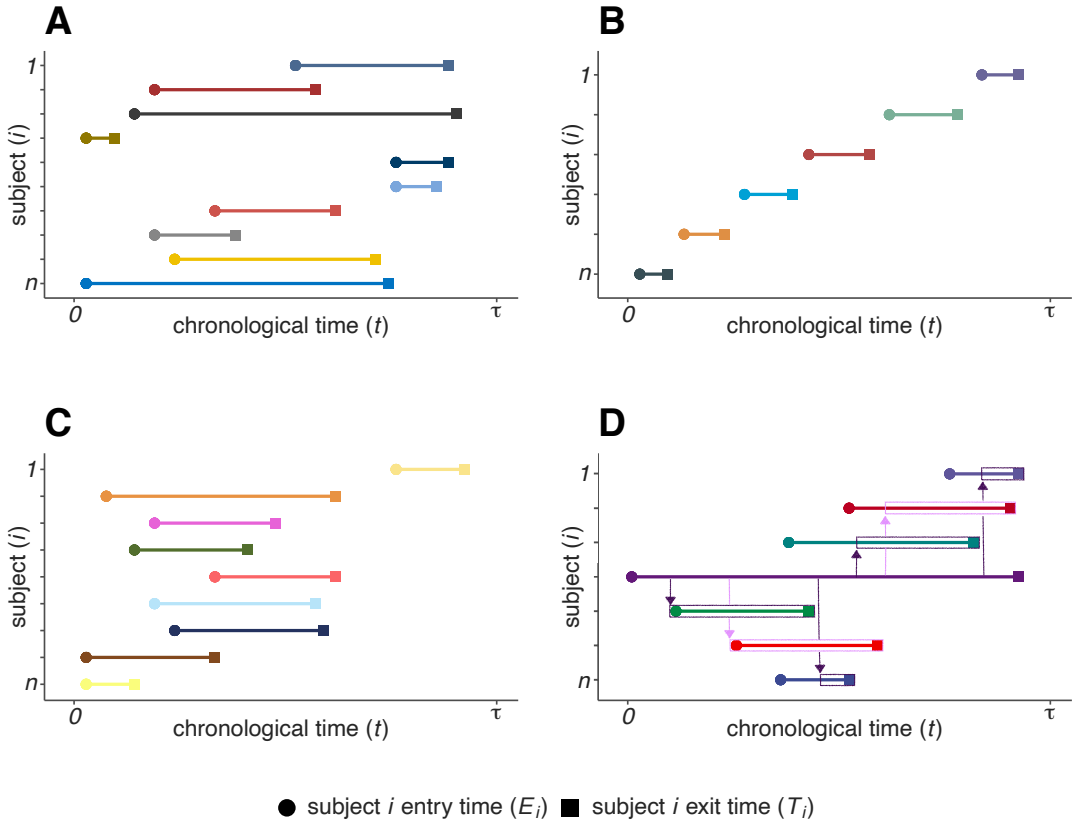}}
  \caption{\textbf{Examples of dynamically streaming time-series data.} The following scenarios are shown: settings where the subjects $i$ have varying entry times $E_i$, exit times $T_i$, observation periods $M_i$.
  In (A), a classic example of dynamic streams is displayed, in which start and exit times vary across subjects and occasionally overlap. In (B), dynamic streams which consist of completely non-overlapping time-series are displayed: an example of time-series that are observed one at a time (such as patients seen by a doctor). In (C), a dynamic setting with large amounts of historical information relative to recently entered time-series (upper-most subject 1) is shown; in this scenario, training algorithms on historical information might be particularly useful to make forecasts for subject 1's time-series as not much data has been accumulated on this subject. 
  In (D) an example of a network of dynamically streaming time-series is provided; in particular, the network for a single subject (middle line on D) is shown. This subject's network is constrained in that they can only interact with at most two other subjects at any given time, and this restriction is illustrated by the two colors for arrows and boxes that branch from the middle subject's time-series. In general, examples of dynamically streaming networks include processes that spread and evolve, like disease and social networks.}
  \label{fig:dynamic}
\end{figure}

In most practical settings, time-series data exhibit a heterogeneous streaming profile comprised of varied length of the series, and diverse start and exit times. In order to accommodate a manifold of different applications, including dynamic enrollment/exit (collectively referred to as dynamic streams), we extend the formulation of the estimation problem described in Section \ref{sect3}, along with the POSL algorithm, in the following sections. In particular, we redefine the observed data, statistical model, and the target parameter in Section \ref{sect6.1}, taking into account the possibly dynamic and disparate tracking of each collected sample. In Section \ref{sect6.2} we describe the appropriate CV for dynamic streams, redefine the loss, online cross-validated selector, ensemble of candidate estimators, and define a new $m$-specific prediction function for dynamic enrollment streaming settings. In Figure \ref{fig:dynamic} we provide examples of dynamic streams, introducing subject-specific time and its relationship to chronological time, and illustrate various ways in which this heterogeneous streaming profile can evolve.

\subsection{Formulation of the Estimation Problem with Dynamic Streams}\label{sect6.1}

Let \gls{gls_E} be an entry time for each new time-series, corresponding to the chronological time domain $t = 1, \ldots, \tau$, for $i = 1, \ldots, n$. We assume a natural ordering for all $E_i$, with $0 \leq E_1 \leq \ldots \leq E_n$ even if multiple samples enroll around the same time. Our assumption on the strictly monotone increasing entry times follows from the fluid definition of $t$; as we put no restrictions on the time definition, we note that for sufficiently small $t$, no subjects exhibit $E_i = E_{i+1}$ even if enrolling a group of units. 

Suppose each unit $i$ is tracked over \gls{gls_Mi} time points starting at $E_i$, where the final chronological time \gls{gls_T} and duration $M_i$ are within the $(0, \tau)$ range. The one-to-one mapping from $t$-chronological time to $m$-individual time is given by $\gls{gls_hi} = t-E_i$, resulting in $m \in \{0, \ldots, M_i\}$; with that, let $h_i(E_i) = 0$ and $h_i(T_i) = M_i$. The function $h_i$ is subject specific as it depends on individual $i$'s start time $E_i$; writing just $h$ denotes a general function which can take a vector of start times. We define the process on each unit $i$ as $m \mapsto \gls{gls_Oi_ds}$ with $O^{'}_i = (O_i(m): m = 0, \ldots, M_i)$ being the full observed data on subject $i$. Let $O^{'}_i(m) = \gls{gls_Oi_h}$ be the observed data on subject $i$ in chronological time $t$; note that, in order to define $O_i(h_i(t))$, we need the current time $t$ and the subject's entry time, $E_i$. As in Section \ref{sect3}, the time-varying part of the observed data has structure corresponding to $O^{'}_i(m) = (W^{'}_i(m), Y^{'}_i(m))$, equivalently written as $O_i(h_i(t)) = (W_i(h_i(t)), Y_i(h_i(t)))$ in chronological time. Here, $Y_i(h_i(t))$ is a response variable and $W_i(h_i(t))$ is a vector of covariates for subject $i$ at collected point $h_i(t)$ in chronological time $t$. We impose an order constraint where, for all $t$, $W(h_i(t))$ occurs before $Y(h_i(t))$, and define $X_i$ as the vector of baseline covariates collected at entry time $E_i$ for subject $i$. We define the total number of subjects in the study at chronological time $t$ as $\gls{gls_nt} = \sum_{i=1}^n \mathbb{I}(E_i \leq t)$, reflecting all trajectories started before (or at) time $t$. Equivalently, we also let \gls{gls_nmt} denote the number of samples with $m$ points up to $t$, where $n_m(t) = \sum_{i=1}^{n(t)} \sum_{s=1}^{t} \mathbb{I}(h(s) = m)$.
We can represent the observed data coming from dynamic streams as a single time-series through chronological time $t$ by defining a process $F$ such that $F(t) = \gls{gls_Fmt} \equiv \{O_i(h_i(t)): \text{all } i \text{ where }E_i \leq t\}$. 
Then, we have that $(F^{n(0)}(0), \ldots, F^{n(\tau)}(\tau))$ reflects a single time-series we can learn from. We emphasize that, for dynamic streaming settings, $F(t)$ describes a collection of all observed time-series enrolled at or before time point $t$.
Finally, in the previous sections we defined time $t$- and sample $i$-specific history as $\overline{O}_i(t-1) = (O_i(1), \ldots, O_i(t-1))$. For dynamic streams, we let the history of the $i$-th time-series until time $t$ be defined as $\overline{O}_i(h_i(t-1)) = (O_i(h_i(0)), \ldots, O_i(h_i(t-1)))$. We define the complete history for all samples until chronological time $t$ as $\overline{O}(h(t-1))$, which includes all trajectories observed by time $t$.
\begin{equation*}
\overline{O}(h(t-1)) = \bar{F}(t-1) = \{\overline{O}_i(h_i(t-1)): \text{ all } i \ \text{where} \ E_i \leq t\}. 
\end{equation*}
Analogue to Section \ref{sect3}, 
let $Z^{'}_i(m-1) = Z_i(h_i(t-1))$ denote the fixed dimensional summary measure of the form $Z^{'}_i(m-1) = Z_i(h_i(t-1)) = f_i(\overline{O}(h(t-1)) \in \mathbb{R}^k$; note that with this formulation, $Z_i(h_i(t-1))$ could support both time and sample dependence, as discussed in previous sections. We define the estimand as a time $m$ prediction problem for the $i^{\text{th}}$ subject:
\begin{equation}
    \Psi(P_{0})(X_i, Z^{'}_i(m-1), m) = \Psi(P_{0})(X_i, Z_i(h_i(t-1)), h_i(t-1))= E_{P_{0,X_i}}[Y^{'}_i(m) | Z^{'}_i(m-1)], 
\end{equation}
where the expectation on the right hand side is taken w.r.t the conditional distribution $P_{0,X_i}$, and $\Psi(P_{0})(X_i,Z^{'}_i(m-1),m)$ is the prediction function for the $i^{\text{th}}$ subject at time-series time $m$ (equivalent to chronological time $h_i(t)$). In particular, we want to learn $(X,Z^{'}(m-1),m) \mapsto \Psi(P_0)(X,Z^{'}(m-1),m)$, and thereby obtain a prediction function for each unit $i$ that predicts $Y^{'}_i(m)$ with $\Psi(P_0)(X_i,Z^{'}_i(m-1),m)$. Further, let $P_{n,t} \mapsto \hat{\Psi}(P_{n,t})$ represent a mapping from $P_{n,t}$ into a predictive function $\hat{\Psi}(P_{n,t})$. We define $\hat{\Psi}(P_{n,t})(X_i, Z^{'}_i(m-1),m)$ as an estimator of $\Psi(P_{0})(X_i, Z^{'}_i(m-1), m)$, and write $\hat{\Psi}(P_{n,t})(X_i,Z^{'}_i(m-1),m) = \psi_{n,t}(X_i,Z^{'}_i(m-1),m)$ as the predicted outcome for unit $i$ of the estimator $\hat{\Psi}(P_{n,t})$ based on $(X_i,Z^{'}_i(m-1),m)$.

\subsection{Personalized Online Super Learner for Dynamic Streams}\label{sect6.2}

Let $C_h(i,t,\cdot) = \gls{gls_Ch} = (X_i,Y_i(h_i(t)),Z_i(h_i(t-1)),E_i,T_i)$ denote the subject $i$ and chronological time $t$ observed data, analogue to Section \ref{sect4}. As previously defined, $B_t$ defines a time-specific split vector such that, for all $1 \leq i \leq n$, $B_t(i, \cdot, E_i,T_i) \in \{-1,0,1\}^t$. Extending the formulation described in Section \ref{sect4} further, we define the following split of the learning set into corresponding three disjoint subsets,
\[
  B_t^v(i,s,E_i,T_i)=\begin{cases}
               -1, \ \  C_h(i,s,E_i,T_i) \  \text{not used}\\
                \ \ \  0, \ \ C_h(i,s,E_i,T_i) \  \text{in the training set}\\
                 \ \ \ 1, \ \ C_h(i,s,E_i,T_i) \  \text{in the validation set,}
            \end{cases}
\]
where our cross-validation scheme now takes into account if we have yet to observe sample $i$ (by chronological time $t$) and how long is its trajectory. Knowing $E_i$ and $T_i$ proves important shortly, as we define how the loss, and the corresponding (online) cross-validated risk are defined and evaluated.

Let $L(\psi)(C_h(i,t,E_i,T_i))$ denote the loss function for the data record $C_h(i,t,E_i,T_i)$ for subject $i$, where $$(X_i, Y_i(h_i(t)), Z_i(h_i(t-1)),E_i,T_i) \mapsto L(\psi)(X_i,Y_i(h_i(t)),Z_i(h_i(t-1)),E_i,T_i).$$ For example, given the prediction function $\psi$, we might want to evaluate its performance using the squared error loss
\begin{align}
L(\psi)(C_h(i,t,E_i,T_i)) = c(i,h_i(t),E_i,T_i)(Y_i(h_i(t)) - \psi(C_h(i,t,E_i,T_i)))^2,
\end{align}
where $c(i,h_i(t),E_i,T_i)$ represents a weight function dependent on sample $i$, time $h_i(t)$ and the unit's entry and exit time. In particular, if $t \leq E_i$ or $t \geq T_i$, we might define $c(i,h_i(t),E_i,T_i) = 0$ resulting in $L(\psi)(C_h(i,t,E_i,T_i)) = 0$. The $c(i,h_i(t),E_i,T_i)$ weight might additionally represent a weight function that down-weights the losses for points $h_i(t)$ for which $n_l(t)$ is small; with that, our prediction function would not be penalized for not having enough data collected up until certain times $l$. 

Let $\gls{gls_I} = \{i: E_i \leq t, t \leq T_i\}$ denote a set of all samples with start date before current chronological time $t$ with data still being collected $(t \leq T_i)$. We define the true risk as the expected value of $L(\psi)(C_h(i,t,E_i,T_i))$  w.r.t the true conditional distribution $P_{0,O_i}$ for each sample $i$:
\begin{align}
R(P_{0}, \psi) &= \sum_{t=1}^{\tau} \sum_{i \in \mathcal{I}}  E_{P_{0,O_i}} [L(\psi)(C_h(i,t,E_i,T_i)) | X_i, Z_i(h_i(t-1)) ]\\
&= \sum_{t=1}^{\tau} \sum_{i \in \mathcal{I}}  E_{P_{0,O_i}} [(Y_i(h_i(t)) - \psi(C_h(i,t,E_i,T_i)))^2 | X_i, Z_i(h_i(t-1))], \nonumber
\end{align}
defined for all subjects that had their start $E_i$ before chronological time $t$, and end date after $t$. Note that, if sample $i$ with start date $E_i \leq t$ also had their end date before time $t$ - then their loss would be undefined, unless an appropriate weighting is part of the loss definition (as discussed above). We note that $R(P_{0}, \psi)$ is an average of all appropriate $i$-specific losses measuring the performance of $\psi$ across all available time points. One might instead be interested in defining a $m$-specific prediction function up until time $\tau$. In particular, the true risk of the $m$-specific prediction function can be written as:
\begin{align}
\gls{gls_risk_m} &= \sum_{t=1}^{\tau} \sum_{i \in \mathcal{I}}  \mathbb{I}(h_i(t) = m) E_{P_{0,O_i}} [L(\psi)(C_h(i,t,E_i,T_i)) |X_i, Z_i(h_i(t-1))]
\end{align}
reflecting an average over all available active samples and times with $m$ time points.

Suppose we have $K$ candidate estimators $\hat{\Psi}_k$, where we denote $\psi_{n,t,k}^0$ as $\psi_{n,t,k}^0 \coloneqq \hat{\Psi}_k(P_{n,t}^0)$ for $k= 1, \ldots, K$. In order to evaluate the time specific performance of each $\hat{\Psi}_k$, we use cross-validation for dependent dynamic steams (which takes into account $E_i$ and $T_i$) in order to estimate the average loss for each candidate $k$ over time. The online cross-validated risk of an online estimator is computed at each time point in chronological time and defined as follows
\begin{align}
R_{CV}(P_{n,t}^1, \hat{\Psi}_k(\cdot)) &= \sum_{j=1}^{t} \sum_{v=1}^V \sum_{(i,s,E_i,T_i) \in \mathcal{B}_{j,v}^1} 
L(\hat{\Psi}_k(P_{n,j}^0))(C_h(i,s,E_i,T_i)) \\
&= \sum_{j=1}^{t} \sum_{v=1}^V \sum_{(i,s,E_i,T_i) \in \mathcal{B}_{j,v}^1}
c(i,h_i(s),E_i,T_i)(Y_i(m_i(s))
- \hat{\Psi}_k(P_{n,j}^0)(C_h(i,s,E_i,T_i)))^2, \nonumber
\end{align}
with mean squared error as the loss. Similarly, one could define the online cross-validated risk \gls{gls_risk_cv_m} of $\hat{\Psi}_k$ for evaluating the cross-validated performance of the prediction function $\psi_{n,t,k}^0$ at time $l$ as:
\begin{align}
R_{CV,m}(P_{n,t}^1, \hat{\Psi}_k(\cdot)) &= \sum_{j=1}^t \sum_{v=1}^V \sum_{(i,s,E_i,T_i) \in \mathcal{B}_{j,v}^1} \mathbb{I}(h_i(s) = m) L(\hat{\Psi}_k(P_{n,j}^0))(C_h(i,s,E_i,T_i))
\end{align}
We define the total online cross-validated risk of $m$-specific prediction functions as $m$-specific risks, with:
\begin{align}
R_{CV}(P_{n,t}^1, \hat{\Psi}_k(\cdot)) = \sum_{m} R_{CV,m}(P_{n,t}^1, \hat{\Psi}_k(\cdot)).
\end{align}
The online cross-validated risk $R_{CV}(P_{n,t}^1, \hat{\Psi}_k(\cdot))$ gives an empirical measure of performance for candidate estimator $k$ trained on training data until chronological time $t$. We define the time $t$ discrete online cross-validation selector as:
\begin{align}
k_{n,t} = \argminD_{k=1,\ldots,K} R_{CV}(P_{n,t}^1, \hat{\Psi}_k(\cdot)),
\end{align}
reflecting the discrete online super learner for all $m$. Instead, we could define a separate selector for the different time points $m$, with the discrete online super learner stratifying the selector by $m$, 
\begin{align}
\gls{gls_kntm} = \argminD_{k=1,\ldots,K} R_{CV,m}(P_{n,t}^1, \hat{\Psi}_k(\cdot)).
\end{align}

Finally, we consider a more flexible online learner that generates a weighted combination of candidate estimators at each time point. Let $\hat{\Psi}_{\alpha}$ be a function of empirical distribution generating an ensemble of $K$ estimators $\{\hat{\Psi}_1, \ldots, \hat{\Psi}_K\}$ indexed by a vector of coefficients $\alpha$. Let $\mathcal{H}$ define a class of weight functions, where $\alpha = \alpha(X) = (\alpha_1(X), \ldots, \alpha_K(X))$ is a collection of $K$ weights that might depend on the baseline covariates $X$, with $\sum_{k=1}^K \alpha_k = 1$ and $\forall \alpha_k$, $\alpha_k \geq 0$. Let $\hat{\Psi}_{\alpha} = \sum_{k=1}^K \alpha_k \hat{\Psi}_{k}$, so that the predictive function based on the training set $P_{n,t}^0$ is given by $\psi_{n,t,\alpha}^0 \coloneqq \sum_{k=1}^K \alpha_k \hat{\Psi}_{k}(P_{n,t}^0)(C_h(i,t,E_i,T_i))$ with $\alpha \in \mathcal{H}$.
We define a $\mathcal{H}$-specific online cross-validation selector for the ensemble as:
\begin{align}
\alpha_{n,t} &= \argminD_{\alpha \in \mathcal{H}} R_{CV}(P_{n,t}^1, \hat{\Psi}_{\alpha}(\cdot)) \\
&= \argminD_{\alpha \in \mathcal{H}} \sum_{j=1}^{t} \sum_{v=1}^V \sum_{(i,s,E_i,T_i) \in \mathcal{B}_{j,v}^1} 
(Y_i(h_i(t)) - \hat{\Psi}_{\alpha}(P_{n,t}^0)(C_h(i,t,E_i,T_i)))^2, \nonumber
\end{align} 
where the loss is defined as the mean squared error. Alternatively, we could compute an online cross-validation selector $\alpha_{n,t,m}$ for each $m$, where
\begin{align}
\gls{gls_alpha_ntm} &= \argminD_{\alpha \in \mathcal{H}} R_{CV,m}(P_{n,t}^1, \hat{\Psi}_{\alpha}(\cdot)).
\end{align} 

\section{Simulations}\label{sect7}
We evaluate the Personalized Online Super Learner implementation described in  Algorithm \ref{posl}, testing its performance and pattern of adaptivity over time for several common time-series settings. For all scenarios, we simulate a total of $31$ time-series with $\tau = 540$ time-points, and repeat the entire procedure $50$ times for a total of $1550$ trajectories. We use a random sample of $30$ time-series to train the Historical learner, and the remaining random sample for the Individual learners. For all simulations described below, we use the same library consisting of a grid of \textsc{xgboost} and \textsc{glm} learners for the Historical and Individual learners implemented in \textsc{sl3} \textsc{R} package \citep{chen2016R, coyle2021sl3, R}. The Historical Super Learner is fit once on the data pooled across individuals whereas the Individual learners, and with it the POSL, are updated every $20$ time points resulting in possibly different fits and weights at different times. In particular, we train sequentially for sample sizes $\{10, 30, 50, \ldots, 470, 490, 510\}$, and evaluate the loss over the last 5 time points of the time-series for which we forecast. To report the performance, we present an average sum over $50$ simulations of convex non-negative least square weights given to $K_H$ Historical and $K_I$ Individual candidate learners over time in Figure \ref{fig:weights}. In addition, we compare the performance of the POSL to an online and non-online ensemble methods using the same library of algorithms, data and test set. In particular, we compare POSL to the canonical online Super Learner algorithm described by \cite{benkeser2018} and the $V$-fold cross-validated Super Learner presented by \cite{sl2007}; by $V$-fold, we refer to $V$-fold cross-validation and the original Super Learner for i.i.d. data. 
The online Super learner was trained using all the samples in a sequential manner: the loss is evaluated over a future five time-point window not seen by the learners, then the fit is updated with new data. The evolution of the mean squared error is shown in Figure \ref{fig:mse}.  

\subsection{Simulation 1: ARIMA Processes}
We start with a simple scenario, making sure that the POSL can learn the conditional mean under the correct data-generating distribution over time for sample $i$ we are interested in. In particular, in Simulation 1, Historical and Individual data are sampled from different data-generating distributions. We sample $30$ time-series from a fifth-order autoregressive integrated moving average process, ARIMA(5,0,0), and a single time-series $i$ from ARIMA(0,0,5). We are interested in optimizing predictions for the single ARIMA(0,0,5) time-series. 

From Figure \ref{fig:weights} (A), we can see that the POSL gives more weight to the Historical learners in the beginning, as there is not enough time points to learn just from the sample $i$. As we collect more data, and the time-series progresses, POSL quickly starts to consistently give more weight to the Individual learners. The Personalized Online Super Learner demonstrates good forecasting performance in terms of the mean squared error at all training times, as shown in Figure \ref{fig:mse} (A), with V-fold Super Learner as a close second.   

\subsection{Simulation 2: ARIMA Processes with \texorpdfstring{$X$}{X}-dependent Common Offset}
In Simulation 2, we build on Simulation 1 by adding a common component to the Historical and Individual data-generating distributions, in order to investigate the performance and behavior of the POSL algorithm in situations where there is significant ``overlap'' in the  data-generating process, but the Individual distribution is different enough that, asymptotically, one would expect the POSL to give more weights to the Individual learner as $t$ gets larger.  

We simulate baseline covariates $X = \{W_1, W_2, W_3\}$ with
\begin{align*}
W_1 &\sim \text{Binomial}(0.5), \\
W_2 &\sim \text{Uniform}(19, 90), \\
W_3 &\sim \text{Uniform}(0, 2).
\end{align*}
We define a $X$-dependent offset as a function of $W_1$, $W_2$ and $W_3$ with $f(X_i) = 0.5W_{1,i} + 0.02W_{2,i} + 0.5W_{3,i}$ being the offset for sample $i$. We sample $30$ time-series from $f(X)$ + ARIMA(5,0,0) process, reflecting the Historical data-generating distribution. We generate sample $i$ from $f(X_i)$ + ARIMA(0,0,5), so the trajectory evolves as a MA process with offset $f(X_i)$. Figure \ref{fig:weights} (B) shows the evolution of weights for the Historical and Individual learners over time. As seen in Simulation 1, the POSL gives more weight to the Historical fit in  beginning, due to the scarce number of time points collected for time-series $i$. As data becomes more abundant, we can characterize the conditional mean for sample $i$ better, giving more weight to the Individual learners. However, from Figure \ref{fig:weights} (B), we can also see that due to the common offset, POSL does not start giving more weight to the Individual fit until about 100 time points are included in the training set --- much further than seen in Simulation 1. As shown in Figure \ref{fig:mse} (B), POSL demonstrates uniformly the best forecasting performance in terms of the mean squared error at all training times.

\subsection{Simulation 3: Interrupted ARIMA Processes}
\begin{figure}[H]
  \includegraphics[scale=0.47]{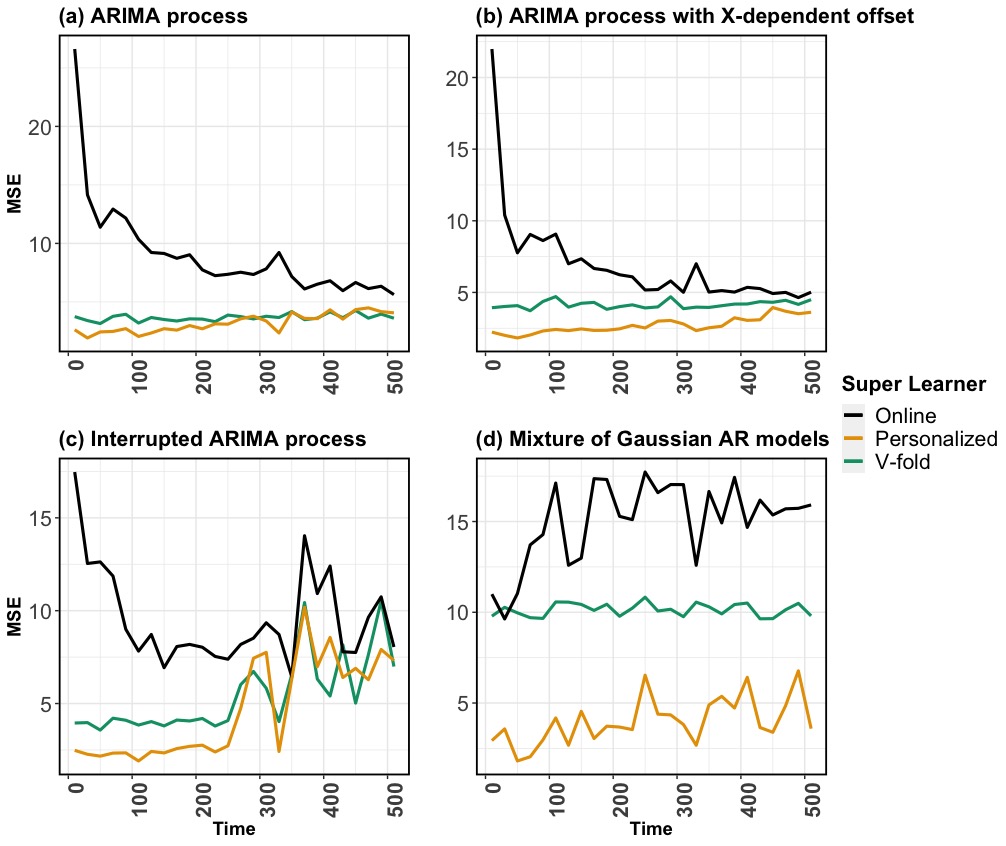}
  \caption{\textbf{MSE over time for different Super Learners.} Evolution over time of the mean squared error for the Personalized Online Super Learner, Online Super Learner and V-fold cross-validated Super Learner. The V-fold cross-validated Super Learner was trained once on the data, whereas Personalized Online Super Learner and Online Super Learner were trained in an online fashion. The loss is evaluated over a five time-point window not seen by the learners for
  (A) Simulation 1, representing different ARIMA processes; (B) Simulation 2, reflecting different ARIMA processes with common baseline covariate dependent offset; (C) Simulation 3, where the time-series is an interrupted ARIMA process; (D) Simulation 4, representing a mixture of Gaussian AR models.} 
  \label{fig:mse}
\end{figure}

We continue to build on Simulation 2 by generating an interrupted time-series as the data-generating distribution for sample $i$ we want to create forecasts for. The motivation for Simulation 3 is to test if POSL can detect changes in the underlying stream of data, and adjust accordingly. We sample $30$ time-series from $f(X)$ + ARIMA(5,0,0) process, representing the Historical data-generating distribution. As in the previous section, we define $f(X)$ as a $X$-dependent function with $f(X_i) = 0.5W_{1,i} + 0.02W_{2,i} + 0.5W_{3,i}$ characterizing the sample $i$ offset. In contrast, we sample trajectory $i$ as an interrupted time-series --- with the first half drawn from $f(X)$ + ARIMA(0,0,5) process, and the second half drawn from the same process as the Historical data-generating distribution, $f(X)$ + ARIMA(5,0,0). From Figure \ref{fig:weights} (C) we can see that the POSL is able to detect changes in the time-series data for sample $i$ as time progresses and more data is collected. As in Figure \ref{fig:weights} (B), the POSL starts with giving more weight to the Historical learners (until about $t=100$), but quickly learns to start giving more weight to the Individual learners as more data is collected. At time $t=270$, at roughly about half the training time, we can see that POSL responds to change in the data-generating process by giving more weight to the Historical learners again --- which is the correct data-generating process for the second half of time-series $i$. The distribution of weights assigned to the Individual learners continues to decrease until the end of training, as demonstrated in Figure \ref{fig:weights} (C), showing that POSL is able to quickly adapt to changes in time-series as time progresses. In Figure \ref{fig:mse} (C) we can see that POSL outperforms all other tested algorithms, except at rare points in the later part of the time-series $i$ when $V$-fold Super Learner slightly outperforms (or performs as well) as POSL; this can be explained by the fact that $V$-fold Super Learner fit is trained only on the samples sampled from $f(X)$ + ARIMA(5,0,0) process, and POSL has to learn the correct current form with a slight delay due to the small batch sizes used for training. 

\subsection{Simulation 4: Finite Mixture of Gaussian Autoregressive Models}

\begin{figure}[H]
  \includegraphics[scale=0.46]{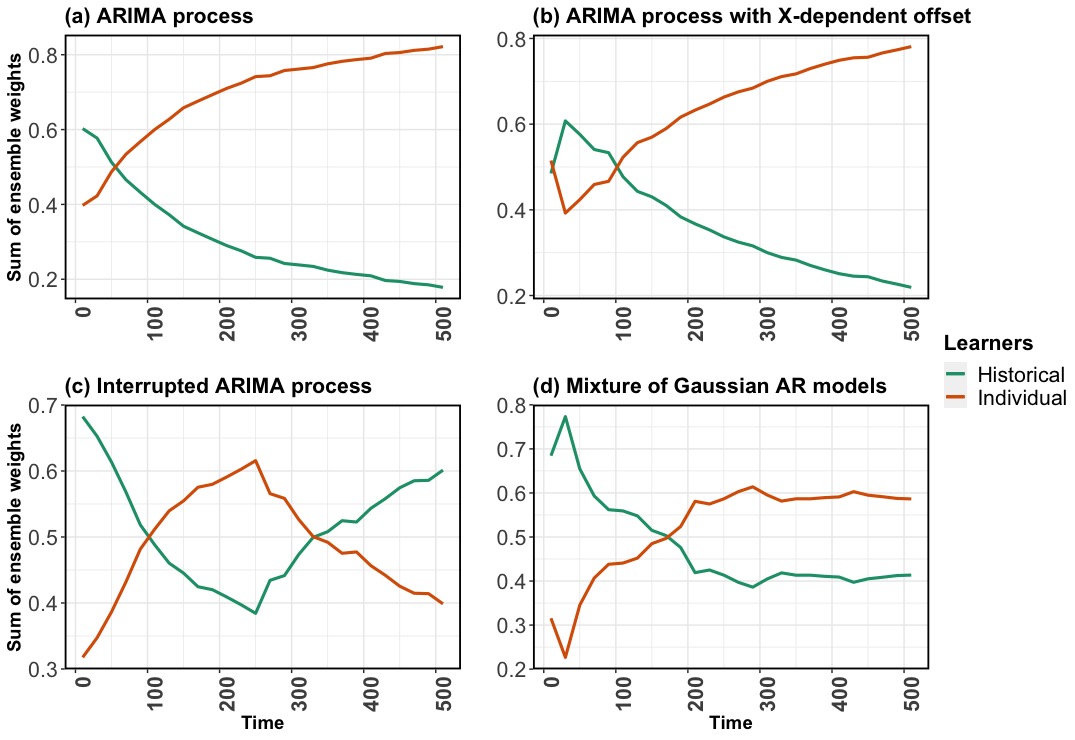}
  \caption{\textbf{Sum of weights over time for Historical and Individual Learners.} Evolution over time of the sum of convex non-negative least square weights for Historical and Individual learners at different training times for (A) Simulation 1, representing different ARIMA processes; (B) Simulation 2, reflecting different ARIMA processes with common baseline covariate dependent offset; (C) Simulation 3, where the time-series is an interrupted ARIMA process; (D) Simulation 4, representing a mixture of Gaussian AR models.} 
  \label{fig:weights}
\end{figure}

In simulation 4, we simulate sets of time-series using the \textsc{GRATIS} package, developed in order to expedite simulation of dependent data with controllable features and provide basis for future time-series benchmarks \citep{kang2020}. The general approach employed is based on Gaussian mixture autoregressive models, used to generate a wide range of non-Gaussian and nonlinear time-series. First developed by \cite{le1996}, mixture transition distribution models used to capture many non-Gaussian and nonlinear features were generalized to Gaussian mixture autoregressive models by \cite{wong2000}. In addition to supporting generation of heterogeneous sets of time-series, \textsc{GRATIS} also provides options for simulating from a random population of mixture autoregressive models with specified features \citep{kang2020}. In particular, we specify common features of the Historical and Individual data-generating distribution including entropy and the smoothed trend component for the Seasonal and Trend decomposition using Loess (STL decomposition). We differentiate the series based on their stability, defined as the variance of non-overlapping window means and the largest mean shift between two consecutive windows. With that, the Historical and Individual series exhibit the same trend and amount of information, but different variance. From Figure \ref{fig:weights} (D), we can see that POSL once again starts with giving more weight to the Historical fit, but eventually switches completely to the Individual learners as more data on the sample in question is collected. In terms of the MSE, POSL shows an uniformly best forecast among all other algorithms across all tested times. 

\section{Data Analysis}\label{sect8}
We illustrate the POSL algorithm in an application for five-minute ahead forecasting of an individual's mean arterial pressure (MAP), which is one of the most important vital signs in the intensive care unit (ICU). Data obtained from the MIMIC II database (Multiparameter Intelligent Monitoring in Intensive Care) included 370 subjects' baseline covariates (age, sex, body mass index, ICU subunit, SAPS II and SOFA mortality scores, ICU admission type), time-varying binary exposures (vasopressors, ventilation, sedation), and time-varying continuous vitals (pulse, heart rate, systolic and diastolic blood pressures, and MAP outcome) \citep{goldberger2000physiobank, saeed2011multiparameter}. Additional covariates were derived from this set of variables, most of them from the time-varying variables, including lagged values of the time-series and summary measures over at most one hour of history. 

A total of 368 subjects were used for training the Historical learners using the \textsc{sl3} R package \citep{coyle2021sl3, R}. The library of Historical learners included the following: multiple variations of gradient boosted decision trees (\textsc{xgboost}), random forests (\textsc{ranger}), and elastic net generalized linear models (\textsc{glmnet}); a discrete Bayesian additive regression trees model  (\textsc{dbarts}), a Bayesian generalized linear model (\textsc{bayesglm}), and a linear regression (\textsc{glm}) \citep{chen2016R, ranger, glmnet, dbarts, arm}. This library was fit after reducing the number of time-varying covariates with a pre-screening step that selected the 200 ``most important'' time-varying covariates according to a \textsc{ranger} random forest variable importance metric, and then those 200 time-varying covariates and the baseline covariates were passed on to the library of Historical learners. The two patients that were not selected for training the Historical learners were used to train, separately, a library of Individual online learners; the selection of the two patients was random. Individual learners were updated with each batch of five observations (i.e., updated every five minutes), following accumulation of an initial training size consisting of 60 observations. The Individual learners included the following: multiple variations of nonlinear time-series models (\textsc{tsDyn}) and elastic net generalized linear models (\textsc{glmnet}); a linear regression (\textsc{glm}), a gradient-boosted decision tree model  (\textsc{xgboost}), a random forest model (\textsc{ranger}), and an ARIMA model with automated tuning (\textsc{auto.arima}) \citep{tsdyn, glmnet, chen2016R, ranger, forecast}. The linear regression and ARIMA Individual learners were fit following a pre-screening step involving lasso regression, in which the variables with non-zero lasso regression coefficients were selected and then passed to these Individual learners.

\begin{figure}[H]
  \makebox[\textwidth][c]{\includegraphics[width=1.12\textwidth]{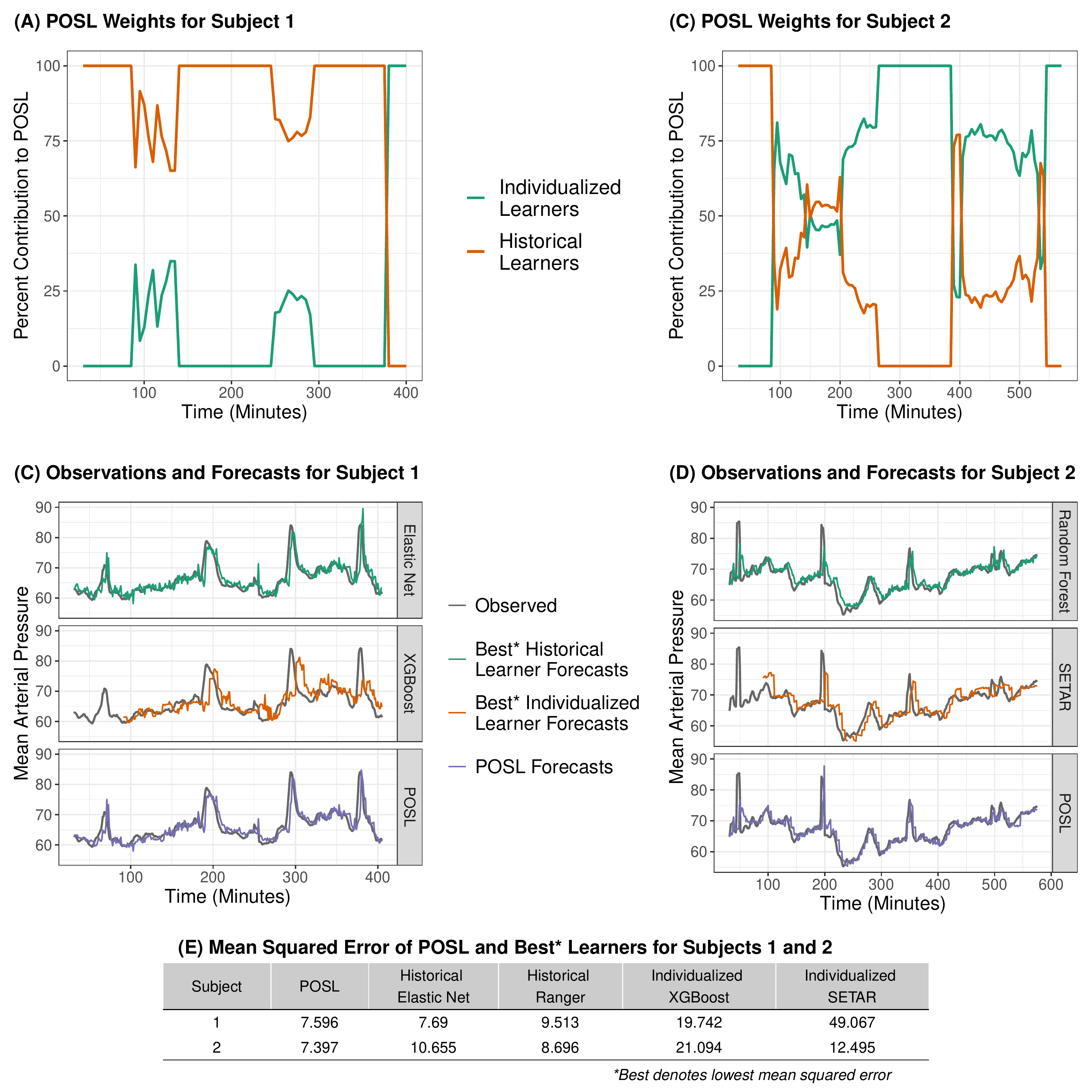}}
  \caption{\textbf{Application of POSL for five-minute ahead forecasting of mean arterial pressure with ICU data}. In (A) and (B), we show how POSL assigned weights over time to the Historical and Individualized candidate learners for subjects 1 and 2, respectively. We present POSL's forecasts and the ``best'' (lowest mean squared error) Historical and Individual learners' forecasts alongside the observed mean arterial pressure in (C) for subject 1, where the best Individualized learner was a gradient boosted regression tree (XGBoost) and the best Historical learner was elastic net regression, and in (D) for subject 2, where the best Individualized learner was a Self Exciting Threshold Autoregressive (SETAR) time-series model and the best Historical learner was random forest. In (E), the mean squared error of all the learners in (C) and (D) is displayed for both subjects.} 
  \label{fig:analysis}
\end{figure}

At each subject-specific 5-minute update, POSL selects the candidate with the lowest online cross-validated risk, where the set of candidates included the Individual and Historical learners, as well as ensembles of them. For both subjects, POSL's risk function was the weighted mean squared error (expectation of the weighted squared error loss), where the weights decreased as a function of time. For losses obtained 180 minutes or more from the subject's current time $m$, the weights were set to 0 when calculating the weighted mean loss. For losses obtained 30 minutes or less from the subject's current time $m$, the weights assigned to those losses were set to 1. The weights assigned to losses that were obtained more than 30 minutes but less than 180 minutes from the subject's current time $m$ decayed as $(1-0.001)^{m-m_L}$, where $m_L$ is the time when the loss was measured, $m_L=0,\ldots,m$, so the difference $m-m_L$ is the lag in time fromloss's time andcurrent time. Let $w(m_L)$ denote the weight assigned to a loss measured at time $m_L$, then this strategy to weight losses based on their lag from the current time can also be expressed as
\[ 
 w(m_L) =
    \begin{cases} 
      0 & \text{ if } m_L \leq m-180 \\
      (1-0.001)^{m-m_L} & \text{ if } m-180 < m_L < m-30 \\
      1 & \text{ if } m_L \geq m-30.
   \end{cases}
\]
In Figure \ref{fig:analysis} we illustrate the application of POSL to the ICU data problem to obtain five-minute ahead forecasts of an individual's mean arterial pressure, summarizing POSL's performance for the two subjects that were not used in training the Historical learners. In parts A and B of Figure \ref{fig:analysis}, we show how POSL assigned weight to the Historical and Individualized candidate learners over time. For each subject, we identified the Individualized learner and the Historical learner that had the lowest MSE when averaged across the individual's time-series, and these ``best'' Individualized and Historical learners varied across the subjects. We present POSL's forecasts and the ``best'' Historical and Individual learners' forecasts alongside the observed mean arterial pressure in parts C and D of Figure \ref{fig:analysis}. We present the mean squared error of all of the learners presented in Figure \ref{fig:analysis} in part E which displays for subject 2, the performance of the learners that performed best for subject 1, and visa versa. This table shows the variability of the performance of the candidate learner's across subjects and the stability of POSL's performance, highlighting POSL's ability to adapt to an individual's time-series by leveraging the candidates and POSL's ability to perform better, in terms of the specified loss, than any of it's candidates.  

\section{Discussion}\label{sect9}

In this manuscript, we consider the problem of generating personalized forecasts in the data streaming setting with multiple time-series of unknown underlying structure. The Personalized Online Super Learner (POSL) is an online ensembling machine learning algorithm which utilizes multiple time-series and ensembling combination methods with the goal of optimizing individual forecasts. The POSL is regularly updated over time using batches of streaming data, and leverages both online pooled (learning across individuals) and individual (learning through time) learners at each time step --- allowing for the ensemble weights to depend on the amount of data collected, stationarity, and level of noise. While relying on both Historical (pooled) and Individual learners, the final forecasts are a product of optimizing predictions for each time-series individually at each time step. The general scenario studied consists of observing $n$ observations comprised of baseline, time-varying and response covariates collected over $\tau$ time points --- possibly with unknown dependence structure among trajectories or trajectories 
sampled from different processes. In addition, we study how the proposed method can be adapted to dynamic enrollment and exit of samples over time. 
We present multiple cross-validation schemes relevant for different streaming settings, and advocate for an adaptive meta-learning step, where the final weights of the ensemble learner are based on mutual characteristics of a group of time-series, or completely individualized. Finally, under stronger conditions then necessary for the setup we describe, 
we apply the results established by \cite{benkeser2018} in a more general time-series setting, with multiple time-series. The established result shows that the performance of the cross-validation based best algorithm is asymptotically equivalent with the performance of the best unknown candidate learner --- providing a powerful way to optimally, and in a personalized way, combine multiple estimators in an online, dependent setting. 

We note that the POSL can be used for estimation of any parameter of the conditional distribution of $O_i$ given its past and past of other, (possibly different) time-series that minimizes the empirical risk. Thinking of all the trajectories as a single ordered time-series provides an interesting opportunity to study the asymptotic behavior of the proposed online super learner in a variety of settings, including dynamic streams. Depending on the number of time points, type of enrollment and dependence across subjects, it is possible to consider asymptotics in time $t$, number of subjects $n$, or a combination. For example, one can study asymptotics in time $t$ only for a fixed number of dependent subjects, or asymptotics in time $t$ for time-series sampled from different data-generating distributions. Alternatively, we could rely on the number of samples only - when subjects are followed up for a limited time frame, when there is no common structure through time, or when the entry times are all concentrated in a finite chronological time interval. For low dependence settings where samples are followed for a long period of time, it is possible to exploit asymptotics in both the total number of time points observed as well as across the $n$ subjects. We emphasize that POSL is able to adapt to the underlying structure in data for all the mentioned settings --- allowing the proposed methodology to pick between relying on structure through time, samples, or both, at each time point. As such, while we impose assumptions on our statistical model in Section \ref{sect3.1} for the sake of obtaining oracle results, our true statistical model does not rely on conditional stationarity in order for POSL to perform well, which is in contrast to the canonical online Super Learner \citep{benkeser2018}. Our proposed method is also constructed to provide optimal forecasts for unit $i$ sampled from $P_{0,O_i}$, instead of a collection of time-series.

Finally, we emphasize that the POSL represents theoretically proven, flexible, open source algorithm for many canonical and custom made time-series prediction problems. While motivated by precision medicine, POSL has a wide range of applications that could be considered, including infectious disease forecasting and financial data. The general algorithm described encompasses varying forecasting horizons, cross-validations, dependence across time, varying enrollment/exit times, tailored ensembling methods, and combinations of individual time-series and pooled algorithms. Our simulation results show superior performance over current state-of-the-art online and ensembling algorithms in terms of MSE across a wide range of forecasting scenarios. In future work, we explore similar formulations of the Personalized Online Super Learner that allow for peak detection and safe updates in possible data drifts.

\acks{The authors would like to thank Jeremy R. Coyle, Ph.D. for all of the valuable comments and insight.}


\newpage

\vskip 0.2in
\bibliography{adapt_sl}

\newpage

\appendix
\section*{Appendix A}
\label{app:theorem}

\noindent
{\bf Lemma 1} {\it The difference between the online cross-validated risk (minimized by $k_{n,t}$) and the online cross-validated true risk (minimized by $\overline{k}_{n,t}$) is a discrete martingale.
}\hfill

\vspace{2mm}
\noindent
{\bf Proof:} 
Let $M_{n}(f) = (R_{CV}(P_{n,t}^1, \hat{\Psi}_k(\cdot)) - R_{CV}(P_{0},\hat{\Psi}_k(\cdot)))$. The difference between centered cross-validated risk $R_{CV}(P_{n,t}^1,\hat{\Psi}_k(\cdot))$ and the true cross-validated risk $R_{CV}(P_{0},\hat{\Psi}_k(\cdot))$ conditional on the filtration defined by the training set is a discrete martingale:
\begin{flalign*}
M_{n}(f) 
& =  \sum_{j=1}^{t} \sum_{(i,s) \in \mathcal{B}_j^1} \  [L(\psi_{n,j,k}^0) - L(\psi_0) (C(i,s))]\\
& \  -  \sum_{j=1}^{t} \sum_{(i,s) \in \mathcal{B}_j^1} \  E_{P_{0,O_i}}[L(\psi_{n,j,k}^0) - L(\psi_0) (C(i,s)) | X_i, Z_i(s-1)] \\
& =  \sum_{j=1}^{t} \sum_{(i,s) \in \mathcal{B}_j^1} \  L(\psi_{n,j,k}^0) (C(i,s)) - E_{P_{0,O_i}}[L(\psi_{n,j,k}^0) (C(i,s)) | X_i, Z_i(s-1)] \\
& =  \sum_{j=1}^{t} \sum_{(i,s) \in \mathcal{B}_j^1} \  f(C(i,s)) - E_{P_{0,O_i}}[f(C(i,s)) | X_i, Z_i(s-1)].
\end{flalign*}

\newpage
\vspace{5mm}
\noindent
{\bf A1.}  {\it There exists a $M_1 < \infty$ for any valid loss function $L$ and $\psi \in \pmb{\Psi}$ such that
\[
\sup_{\psi \in \pmb{\Psi}} \sup_{C(i,s)} |L(\psi)(C(i,s)) - L(\psi_0)(C(i,s))| \leq M_1.
\]
}\hfill

\noindent
{\bf A2.}  {\it There exists a $M_2 < \infty$ for $\psi \in \pmb{\Psi}$ so that with probability 1,
\[
\sup_{\psi \in \pmb{\Psi}} \frac{P_{0,O_i}[L(\psi)-L(\psi_0)]^2}{P_{0,O_i}[L(\psi)-L(\psi_0)]} \leq M_2 < \infty
\]
}\hfill

\noindent
{\bf A3.}  {\it There exists a slowly increasing sequence $M_3 < \infty$ such that with probability tending to 1, we have
\[
\frac{1}{M_3} < \frac{d_{0,t}(\psi_{n,t,k_{n,t}}, \psi_0)}{E_{P_{0,O_i}} [d_{0,t}(\psi_{n, t, k_{n,t}}, \psi_0)]} < M_3
\]
and 
\[
\frac{1}{M_3} < \frac{d_{0,t}(\psi_{n, t, \bar{k}_{n,t}}, \psi_0)}{E_{P_{0,O_i}} [d_{0,t}(\psi_{n, t, \bar{k}_{n,t}}, \psi_0)]} < M_3.
\]
}\hfill

\noindent
{\bf A4.}  {\it Given that $M_3$ is a sequence that grows arbitrarily slow to infinity,
\[
t{M_3}^{-3} \min_k E_{P_{0,O_i}}[d_{0,t}(\psi_{n,t,k}),\psi_0)] \rightarrow \infty
\]
as $t \rightarrow \infty$.
}\hfill

\noindent
{\bf Theorem 1} {\it Let $P_0^n$ describe the true data-generating distribution $P_0^n \in \mathcal{M}$, with the target parameter defined as $\Psi : \mathcal{M} \rightarrow \pmb{\Psi}$ evaluated at a particular $P \in \mathcal{M}$. We establish the cross-validation selector $k_{n,t}$ as the minimizer of the cross-validated risk, and the oracle selector $\overline{k}_{n,t}$ as the minimizer of the true cross-validated risk. Under assumptions A1-A4, there exists a constant $C(M_1) < \infty$ such that:
\[
E_{P_{0,O_i}} [(d_{0,t}(\psi_{n,t,k_{n,t}},\psi_0))] \leq E_{P_{0,O_i}} [(d_{0,t}(\psi_{n,t,\overline{k}_{n,t}},\psi_0))] + C(M_1)[\frac{\log(1 + K(t))}{t}]^{1/2}
\]
}\hfill

\noindent
{\bf Proof}.
Under Lemma 1, the proof is a direct generalization of the oracle inequality for a single time-series proved in \cite{benkeser2018} to multiple time-series under cross-validation schemes described in Section \ref{sect4}, assuming conditional stationarity. 



\newpage

\section*{Appendix B}\label{app:CVS}

\begin{figure}[H]
  \includegraphics[scale=0.45]{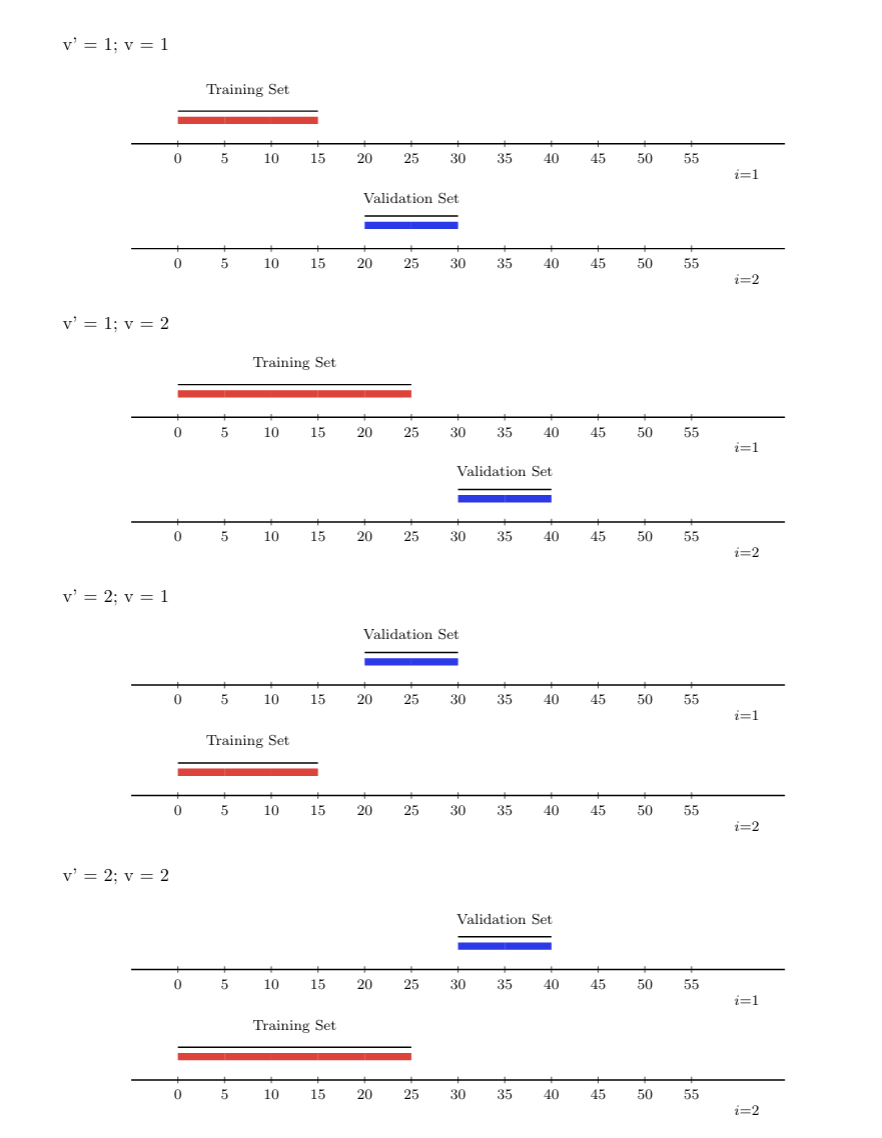}
  \caption{Rolling origin $V$-fold cross-validation illustration $V=2$ v'-wise folds (i.e. sample splitting) and $V=2$ time-series folds (i.e. splitting across time), with initial training set size $n_{t,v}^0=15$, validation set size $n_{t,v}^1=10$, batch size $m=10$, gap $h=5$ and 2 unique id's.}
  \label{fig::rov_cv}
\end{figure}

\begin{figure}[H]
  \includegraphics[scale=0.45]{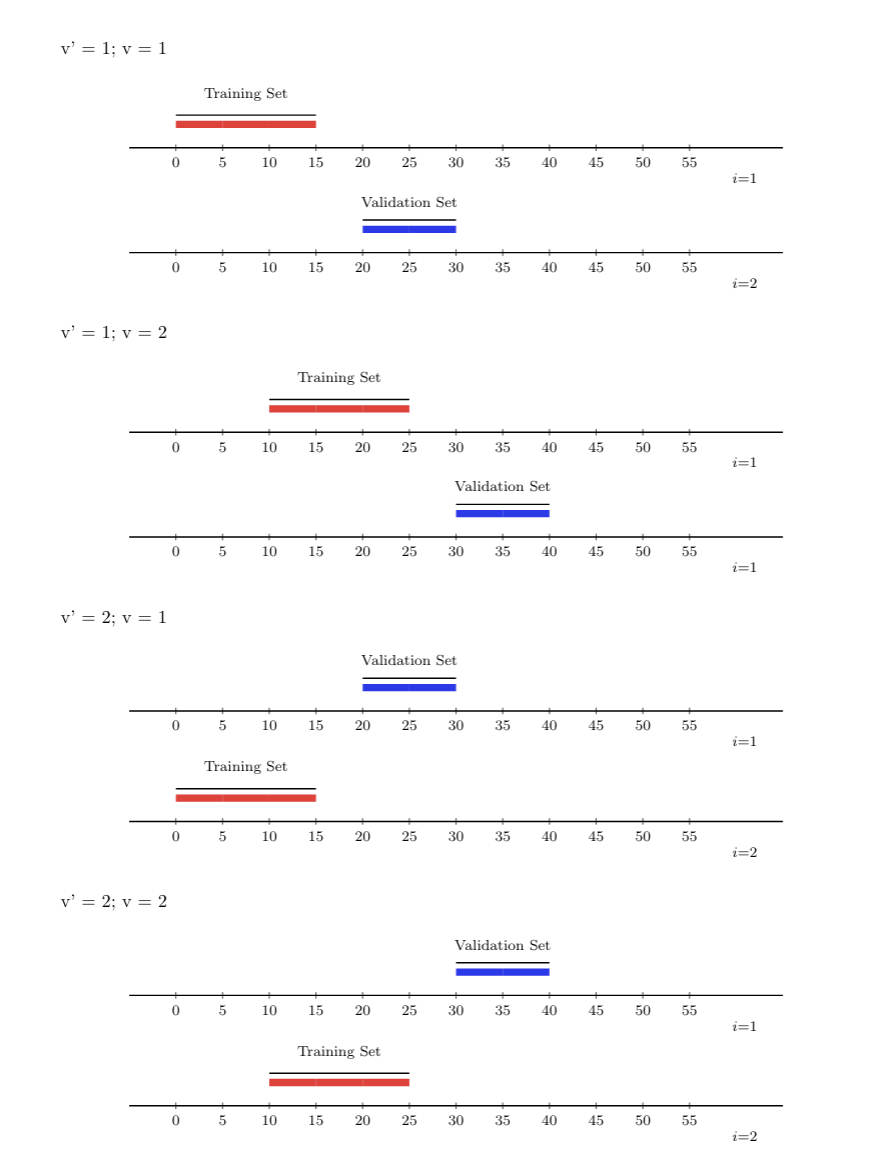}
  \caption{Rolling window cross-validation scheme for $V=2$ v'-wise folds (i.e. sample splitting) and $V=2$ time-series folds (i.e. splitting across time) with $n_{t,v}^0=15$, $n_{t,v}^1=10$, $m=10$, $h=5$ and two unique ids.}
  \label{fig::rwv_cv}
\end{figure}

\newpage
\printnoidxglossary[sort=use, style=altlist]

\end{document}